\RequirePackage{fix-cm}
\documentclass[twocolumn]{svjour3}          
\smartqed  
\usepackage{microtype}
\usepackage{graphicx}
\usepackage{subfigure}
\usepackage{booktabs} 
\usepackage{color}
\usepackage{amsmath,amssymb}
\usepackage{stmaryrd}
\usepackage{multirow}
\usepackage{tabularx}
\usepackage{arydshln}
\usepackage{makecell}
\usepackage{natbib}
\usepackage[colorlinks=true,citecolor=blue]{hyperref}
\graphicspath{{./images/}} 

\newcolumntype{E}{>{\centering\arraybackslash}m{0.75cm}}
\newcolumntype{S}{>{\centering\arraybackslash}m{1cm}}
\newcolumntype{B}{>{\centering\arraybackslash}m{1.6cm}}
\newcolumntype{D}{>{\centering\arraybackslash}m{2.4cm}}

\newcommand{\ie}{\textit{i.e.}}
\newcommand{\eg}{\textit{e.g.}}

\usepackage{hyperref}

\begin{document}

\title{Excitation Dropout:\\ Encouraging Plasticity in Deep Neural Networks
}

\titlerunning{ }        

\author{Andrea Zunino$^\star$\textsuperscript{,}$^\dagger$\textsuperscript{,}$^1$ \and Sarah Adel Bargal$^\star$\textsuperscript{,}$^2$ \and Pietro Morerio$^{1,3}$  \and \\ Jianming Zhang$^{4}$  \and Stan Sclaroff$^{2}$ \and Vittorio Murino$^{1,3,5}$
    \thanks{$^\star$ \rm Equal contribution.}
    \thanks{$^\dagger$ This work was done when A. Zunino was in PAVIS, at Istituto Italiano di Tecnologia.}
}

\authorrunning{ } 

\institute{$^1$ Ireland Research Center, Huawei Technologies Co. Ltd., Dublin, Ireland \\
    $^2$ Department of Computer Science, Boston University, Boston, USA \\
    $^3$ Pattern Analysis \& Computer Vision (PAVIS), Istituto Italiano di Tecnologia (IIT), Genoa, Italy \\
    $^4$ Adobe Research, San Jose, USA \\
    $^5$ Department of Computer Science, University of Verona, Verona, Italy\\
    E-mail: \textit{\{andrea.zunino,pietro.morerio,vittorio.murino\}@iit.it, \{sbargal,sclaroff\}@bu.edu, jianmzha@adobe.com  }}




\date{}

\maketitle

\begin{abstract}
We propose a guided dropout regularizer for deep networks based on the evidence of a network prediction defined as the firing of neurons in specific paths. In this work, we utilize the evidence at each neuron to determine the probability of dropout, rather than dropping out neurons uniformly at random as in standard dropout. In essence, we dropout with higher probability those neurons which contribute more to decision making at training time. This approach penalizes high saliency neurons that are most relevant for model prediction, \ie~those having stronger evidence. By dropping such high-saliency neurons, the network is forced to learn alternative paths in order to maintain loss minimization, resulting in a plasticity-like behavior, a characteristic of human brains too. We demonstrate better generalization ability, an increased utilization of network neurons, and a higher resilience to network compression using several metrics over four image/video recognition benchmarks. Our code is available at \url{https://github.com/andreazuna89/Excitation-Dropout}.
\end{abstract}

\section{Introduction}

Dropout \cite{Hintondrop2012,srivastava2014dropout} is a classical regularization technique that is used in many state-of-the-art deep neural networks, typically applied to fully-connected layers. Standard Dropout selects a fraction of neurons to randomly drop out by zeroing their forward signal. In this work, we propose a scheme for biasing this selection, which utilizes the contribution of neurons to the prediction made by the network at a certain training iteration.

It is well known that dropout avoids overfitting on training data, allowing for better generalization on unseen test data. A recent variant of dropout that targets improved generalization ability is Curriculum Dropout \cite{morerio2017curriculum}: it targets adjusting the dropout rate by exponentially increasing the unit suppression rate during training, answering the question \textit{How many neurons to drop out over time?} Like Standard Dropout \cite{Hintondrop2012,srivastava2014dropout}, Curriculum Dropout selects the neurons to be dropped randomly. In this work, however, we target at determining how the dropped neurons are selected, answering the question \textit{Which neurons to drop out?} 



\begin{figure*}[t!]
\centering
\includegraphics[width=1\linewidth, height = 0.4\linewidth]{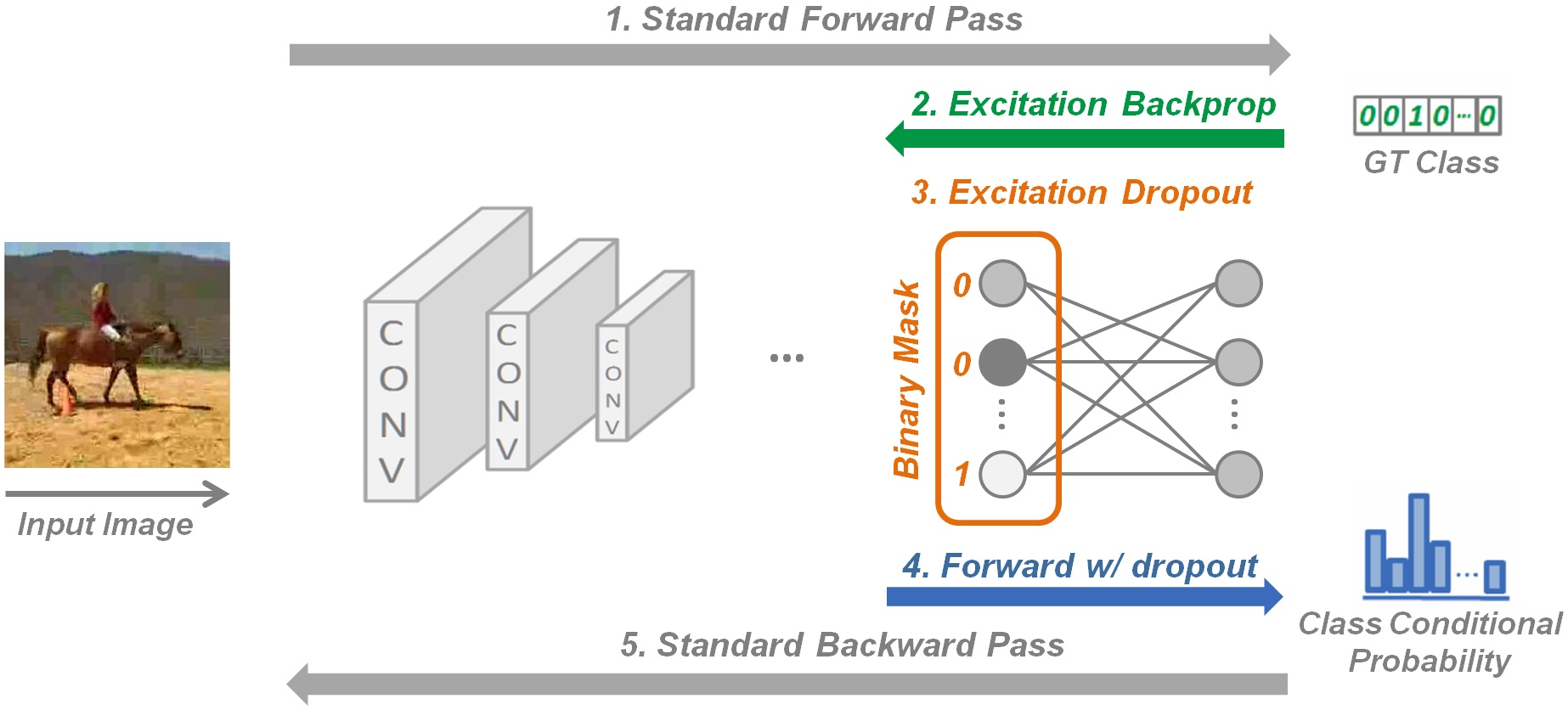}
\caption{Training pipeline of Excitation Dropout. \textit{Step 1:} A minibatch goes through the standard forward pass. \textit{Step 2:} Backward EB is performed until the specified dropout layer; this gives a neuron saliency map at the dropout layer in the form of a probability distribution. \textit{Step 3:} The probability distribution is used to generate a binary mask for each image of the batch based on a Bernoulli distribution determining whether each neuron will be dropped out or not. 
\textit{Step 4:} A forward pass is performed from the specified dropout layer to the end of the network, zeroing the activations of the dropped out neurons. \textit{Step 5:} The standard backward pass is performed to update model weights.}
\label{fig:pipeline}
\end{figure*}

Our approach is inspired by brain plasticity \cite{hebb2005organization,song2000competitive,mittal2018recovering,miconi2018differentiable}. We deliberately, and temporarily, paralyze/injure neurons to enforce learning alternative paths in a deep network. At training time, neurons that are more relevant to the correct prediction are given a higher dropout probability. The relevance of a neuron for making a certain prediction is quantified using Excitation Backprop, a top-down saliency approach proposed by \cite{zhang2016top}. Excitation Backprop conveniently yields a probability distribution at each layer that reflects neuron saliency, or neuron contribution to the prediction being made. This is utilized in the training pipeline of our approach, named \textit{Excitation Dropout}, which is summarized in Fig.~\ref{fig:pipeline}. 

Dropout can be interpreted as a model averaging technique, \ie~an economical approximation using a very large ensemble of networks \cite{Baldi2013}. Formally, consider applying dropout to a single fully-connected layer with $N$ units: there are $2^N$ possible sub-networks that can be sampled. Standard Dropout selects, at each iteration, a random sub-network for training, with probability $p = \binom{N}{n} / 2^N$, which is dependent upon the number of dropped neurons $n$. Conversely, our proposed Excitation Dropout selects with higher probability the sub-networks that least contribute to the correct prediction, \ie~ the worst performing sub-networks, for further training. Selecting, with higher probability, the weaker sub-networks for further training results in a more robust ensemble.


In particular, we first study how this approach improves generalization through an increased utilization of the network's neurons for image classification. We report an increased recognition rate for CNN models that are both fine-tuned and trained from scratch. This improvement is validated on four image/video recognition datasets, and ranges from 1.1\% to 6.9\% over state-of-the-art dropout variants. 

Next, we examine the effect of our approach on network utilization. \cite{mittal2018recovering} and \cite{ma2017less} introduce metrics that measure network utilization. We show a consistent increased network utilization using Excitation Dropout on all datasets considered. For example, averaged over all four benchmarks, we get a 76.55\% reduction in conservative filters, which are the filters whose parameters do not change significantly during training, as compared to Standard Dropout. 

We then study network resilience to neuron dropping at test time. We observe that training with Excitation Dropout leads to models that are a lot more robust when layers are shrunk/compressed by removing units. We demonstrate this when dropping the most relevant neurons, the least relevant neurons, and with a random dropping selection. This can be quite desirable for compressing/distilling \cite{hinton2015distill} a model, \eg~for deployment on mobile devices.

Finally, we propose a lighter version of Excitation Dropout, named \textit{Activation Dropout}. Unlike Excitation Dropout, Activation Dropout does not require a backward pass, and only utilizes the activations of the forward pass to determine the dropout probability. Activation Dropout performs marginally worse than Excitation Dropout but requires less computational overhead.

Effectiveness of regularization methods for training neural networks depends on the architecture and training process, and is still an open problem. For example, batch normalization, another widely used regularization technique, cannot handle small batches. In addition, some architectures  employ multiple regularization techniques, for example Wide-ResNet \cite{zagoruyko2016wide} employs both batch normalization and dropout.

In summary, by encouraging plasticity-like behavior, our novel contributions are threefold:
\begin{enumerate}
\item
\vspace{0.5em}
Better generalization on test data.
\item
\vspace{0.5em}
Higher utilization of network neurons.
\item
\vspace{0.5em}
Resilience to network compression.
\vspace{0.5em}
\end{enumerate}

The rest of the paper is organized as follows. Section~\ref{rel_work} reviews related works on different variants of dropout. Section~\ref{method} introduces our proposed approach, Excitation Dropout. Section~\ref{exps} presents our experimental setup and results on four image/video recognition datasets and on several deep architectures. Section~\ref{act} presents a lighter version of Excitation Dropout, Activation Dropout that we apply on the large-scale ImageNet. Section~\ref{conclus} draws the conclusions.


\section{Related Work}
\label{rel_work}

Dropout was first introduced by \cite{Hintondrop2012} and \cite{srivastava2014dropout} as a way to prevent neural units from co-adapting too much on the training data by randomly omitting subsets of neurons at each iteration of the training phase. 

Some follow-up works have explored different schemes for determining how much dropout is applied to neurons/weights.
\cite{wager2013dropout} described the dropout mechanism in terms of an adaptive regularization, establishing a connection to the AdaGrad algorithm. Inspired by information theoretic principles,  \cite{achille2018information} propose Information Dropout, a generalization dropout which can be automatically adapted to the data. \cite{kingma2015variational} showed that a relationship between dropout and Bayesian inference can be extended when the dropout rates are directly learned from the data. \cite{kang2017shakeout} introduces Shakeout which instead of randomly discarding units as dropout does, it randomly enhances or reverses each unit’s contribution to the next layer. \cite{gal2017concrete} proposed an approach to tune the dropout probabilities using gradient methods. \cite{wan2013regularization} introduced the DropConnect framework, adding dynamic sparsity on the weights of a deep model. DropConnect generalized Standard Dropout by randomly dropping the weights rather than the neuron activations in the network. \cite{rennie2014annealed} proposed a time scheduling for the retaining probability for the neurons in the network. The presented adaptive regularization scheme smoothly decreased in time the number of neurons turned off during training. Recently, \cite{morerio2017curriculum}  proposed Curriculum Dropout to adjust the dropout rate in the opposite direction, exponentially increasing unit suppression rate during training, leading to a better generalization on unseen data. Differently to the previous cited works, our approach does not determine how much dropout is applied.

Other works focus on which neurons to drop out. Dropout is usually applied to fully-connected layers of a deep network. Conversely, \cite{ghiasi2018dropblock} proposed dropping units in a contiguous region of convolutional feature maps. \cite{wu2015towards} studied the effect of dropout in convolutional and pooling layers. The selection of neurons to drop depends on the layer where they reside.  In contrast, we select neurons within a layer based on their contribution.  \cite{wang2013fast} demonstrate that sampling neurons from a Gaussian approximation gave an order of magnitude speedup and more stability during training. \cite{gomez2018targeted} proposed a train time dropout strategy for post hoc pruning of network weights.  \cite{li2016improved} proposed to use multinomial sampling for dropout, \ie~keeping neurons according to a multinomial distribution with specific probabilities for different neurons. \cite{ba2013adaptive} proposed Adaptive Dropout; jointly training a binary belief network with a neural network to regularize its hidden units by selectively setting activations to zero accordingly to their magnitude. While this takes into consideration the magnitude of the forward activations, it does not take into consideration the relationship of these activations to the ground-truth. In contrast, we drop neurons based on how they contribute to a network's decision. To the best of our knowledge, we are the first to probabilistically select neurons to dropout based on their task relevance.


\begin{figure}[t!]
\centering
\includegraphics[width=1\linewidth,height=0.6\linewidth,trim={0cm 1.5cm 0cm 2.3cm},clip]{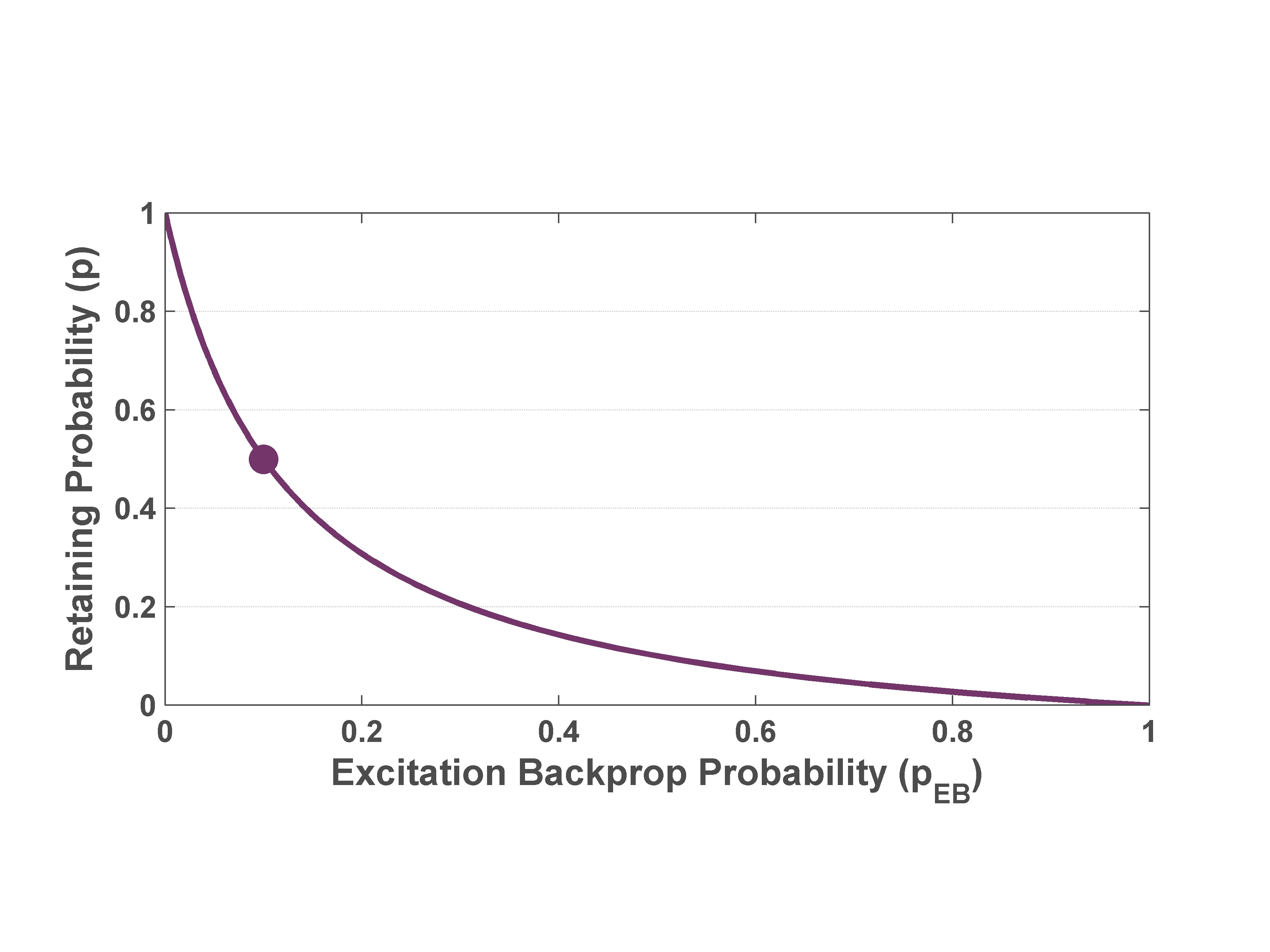}
\caption{The retaining probability, $p$, as a function of the Excitation Backprop probability $p_{EB}$. This plot was created using $N=10$ and a base retaining probability $P=0.5$. In this case, when the saliency of neurons is uniform, \ie~$p_{EB} = 0.1$, then $p=P$ as marked in the figure.}
\label{fig:prob_retain}
\end{figure}

\section{Method}
\label{method}
\subsection{Background}


Saliency maps that quantize the importance of class-specific neurons for an input image are instrumental to our proposed scheme. Popular approaches include Class Activation Maps (CAM) \cite{cam}, Gradient-weighted Class Activation
Mapping (Grad-CAM) \cite{gradcam}, and Excitation Backprop (EB) \cite{zhang2017top}. A thorough analysis of all saliency methods is out of the scope of this work, and the saliency problem in general is far from solved. We choose to use EB since it produces a valid probability distribution for each network layer. The saliency maps obtained using this approach are evaluated for spatial localization of objects and demonstrate the ability of pointing to the right region of an image \cite{zhang2017top}. 

In a standard CNN, the forward activation of neuron $a_j$  is computed by $\widehat{a}_j=\phi(\sum_iw_{ij}\widehat{a}_i+b_i)$, where $\widehat{a}_i$ is the activation coming from the previous layer, $\phi$ is a nonlinear activation function, $w_{ij}$ and $b_i$ are the weight from neuron $i$ to neuron $j$ and the added bias at layer $i$, respectively. 
EB devises a backpropagation formulation that is able to reconstruct the evidence used by a deep model
to make decisions. It computes the probability of each neuron recursively using conditional probabilities $P(a_i|a_j)$ in a top-down order starting from a probability distribution over the output units, as follows:
\begin{equation}
\label{eq:EBmain}
P(a_i) =\sum_{a_j\in \mathcal{P}_i}P(a_i|a_j)P(a_j)
\end{equation}
where $\mathcal{P}_i$ is the parent node set of $a_i$. EB passes top-down signals through excitatory connections having non-negative activations, excluding from the competition inhibitory ones. EB is designed with an assumption of non-negative activations. Most modern CNNs use ReLU activation functions, which satisfy this assumption. Therefore, negative weights can be assumed to not positively contribute to the final prediction. 
Assuming $C_j$ the child node set of $a_j$, for each
$a_i \in C_j$, the conditional winning probability $P(a_i |a_j)$ is defined as
\begin{equation}
P(a_i|a_j) = 
\begin{cases}
    Z_j \widehat{a}_i w_{ij},     & \text{if } w_{ij}\geq 0,     \\
    0,                              & \text{otherwise}
    \end{cases}
    \end{equation}
   where $Z_j$ is a normalization factor such that a probability distribution is maintained, \ie~$\sum_{a_i\in \mathcal{C}_j}P(a_i|a_j) = 1$.
Recursively propagating the top-down signal and preserving the sum of backpropagated probabilities, it is possible to highlight the salient neurons in each layer using Eqn.~\ref{eq:EBmain}, \ie~neurons that mostly contribute to a specific task. We will refer to the distribution of $P(a_i)$ as $p_{EB}(a_i)$.

\subsection{Excitation Dropout}
In the standard formulation of dropout \cite{Hintondrop2012,srivastava2014dropout}, the suppression of a neuron in a given layer is modeled by a
Bernoulli random variable $p$ which is defined as the probability of retaining a neuron, $0 < p \leq 1$. Given a specific layer where dropout is applied, during the training phase, each neuron is turned off with a probability $1-p$.

We argue for a different approach that is \textit{guided} in the way it selects neurons to be dropped. In a training iteration, certain paths have high excitation contributing to the resulting classification, while other regions of the network have low responses. We encourage the learning of alternative paths (plasticity) through the temporary damaging of the currently highly excited path. We re-define the probability of retaining a neuron as a function of its contribution in the currently highly excited path
\begin{equation}
\label{eq:EBdrop}
p=1-\frac{(1-P)*(N-1)*p_{EB}}{((1-P)*N-1)*p_{EB}+P}
\end{equation}
where $p_{EB}$ is the probability backpropagated through the EB formulation (Eqn.~\ref{eq:EBmain}) in layer $l$, $P$ is the \textit{base} probability of retaining a neuron when all neurons are equally contributing to the prediction and $N$ is the number of neurons in a fully-connected layer $l$ or the number of filters in a convolutional layer $l$. The retaining probability defined in Eqn.~\ref{eq:EBdrop} drops the neurons that contribute the most to the recognition of a specific class, with higher probability. Dropping out highly relevant neurons, we retain less relevant ones and thus encourage them to awaken. 

\begin{table}[t]
	\centering
	\begin{tabular}{c|c|c|c}
		\hline
		\textbf{Type} & \textbf{Layer Size}    & \textbf{Filter Size} & \textbf{Pad./Stride} \\ \hline
		conv       & 96 filters    & 5x5         & 2/1            \\ 
		max pool   &               & 3x3         & 0/2            \\ 
		conv       & 128 filters    & 5x5         & 2/1            \\
		max pool   &               & 3x3         & 0/2            \\ 
		conv       & 256 filters   & 5x5         & 2/1            \\ 
		max pool   &               & 3x3         & 0/2            \\ 
		fc         & 2048 units    &             &                \\ 
		fc         & 2048 units    &             &                \\ 
		softmax    & \# classes &             &                \\ \hline
	\end{tabular}
	\caption{Details of the CNN-2 architecture used for experiments on the Cifar10, Cifar100, and Caltech-256 datasets. }
	\label{tab:CNN2}
\end{table}

\begin{table*}[t]
\centering
\begin{tabular}{l|c|c|c|c|c} 
\hline
\textbf{Dataset} & \textbf{Architecture} & \makecell{\textbf{Adaptive} \\ \textbf{Dropout (\%)}}  & \makecell{\textbf{Information} \\ \textbf{Dropout (\%)}} & \makecell{\textbf{Curriculum} \\ \textbf{Dropout (\%)}} & \makecell{\textbf{Excitation} \\ \textbf{Dropout (\%)}} \\ 
\hline
\textit{\textbf{Cifar10}} & CNN-2 & 76.82  &79.41 & 80.03 & \bf{81.94} \\
\textit{\textbf{Cifar100}} & CNN-2 & 44.55  & 49.79 &50.74 & \bf{52.04} \\
\textit{\textbf{Caltech256}} & CNN-2 & 23.32  & 28.66 &28.91 & \bf{35.77} \\
\textit{\textbf{UCF101}}  & AlexNet & 63.96 & 64.21& 64.55 & \bf{67.56} \\
\hline
\end{tabular}
\vspace{0.3em}
\caption{Accuracy comparison for: Adaptive Dropout, Information Dropout, Curriculum Dropout, and Excitation Dropout. Excitation Dropout has highest classification accuracy results compared to the other dropout variants on the four benchmark datasets: Cifar10, Cifar100, Caltech256, and UCF101. The numbers reported in this table are the average test set accuracy over five trained models for each dataset.}
\label{table:analysis}
\end{table*}


\begin{figure*}[t!]
    \centering
        \includegraphics[width=0.4\textwidth,trim={0cm 0cm 0.6cm 0cm},clip]{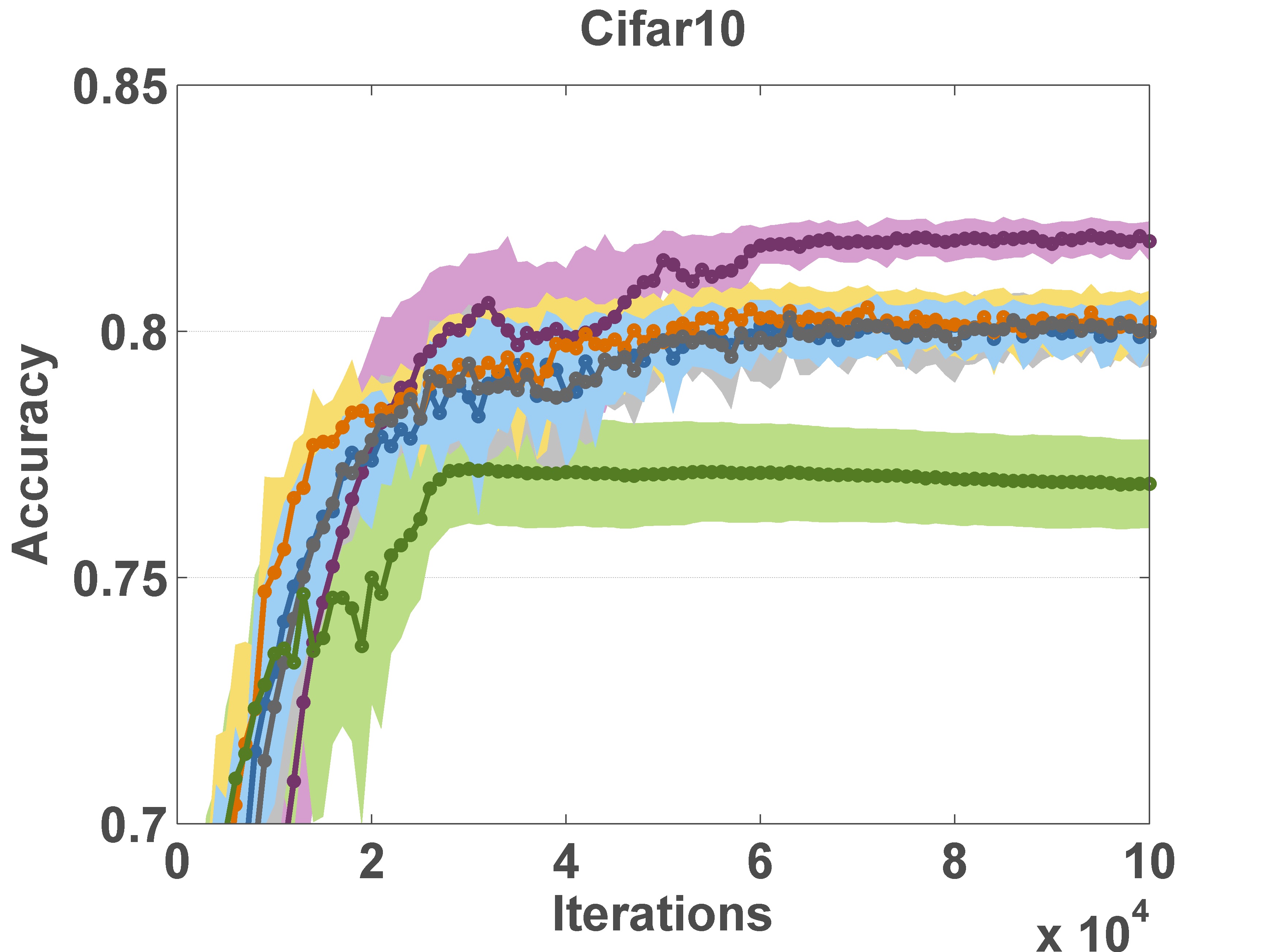}
    ~ 
        \includegraphics[width=0.4\textwidth,trim={0cm 0cm 0.5cm 0cm},clip]{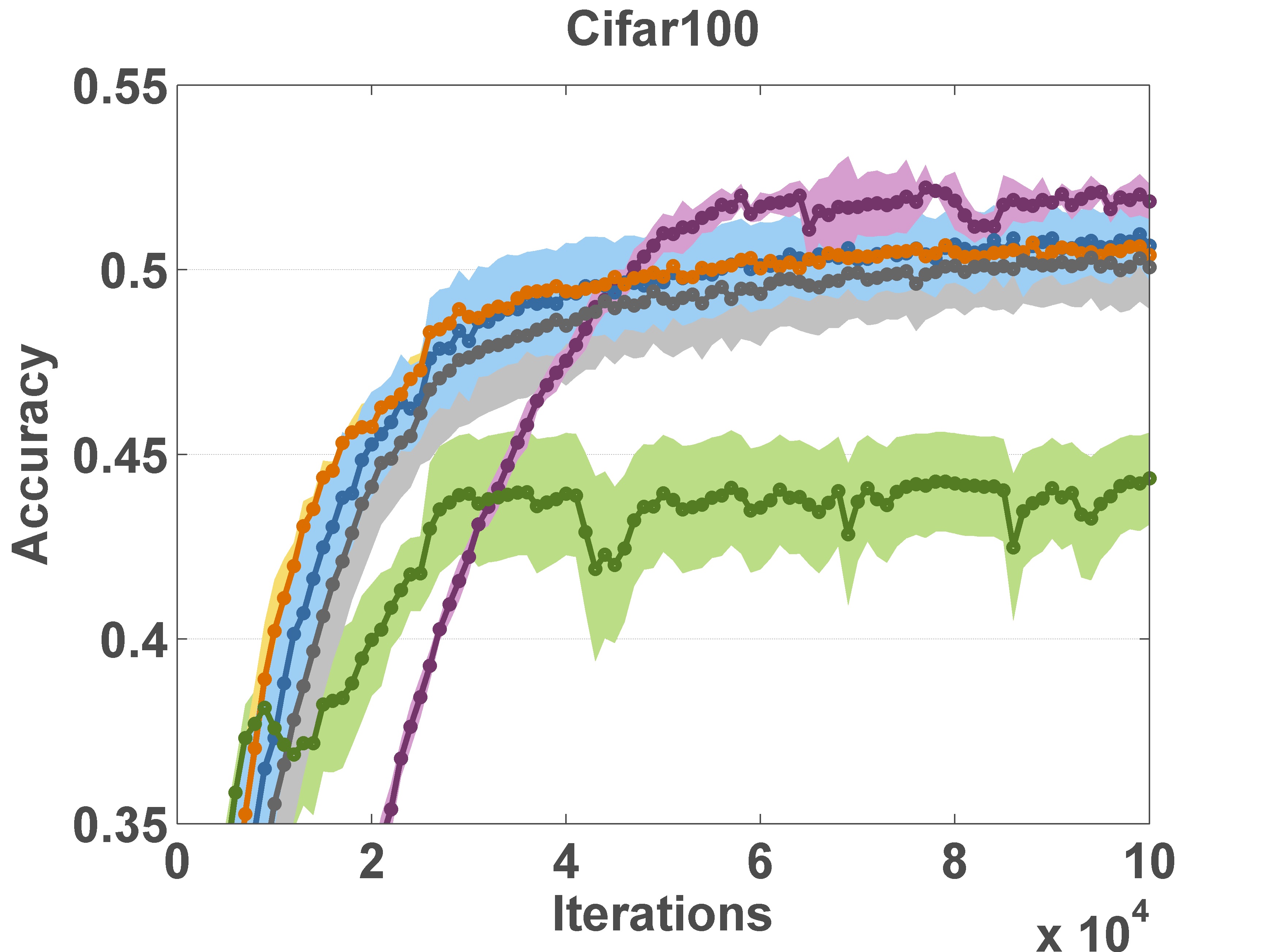}\\[2.5em]
        \includegraphics[width=0.4\textwidth,trim={0cm 0cm 0.6cm 0cm},clip]{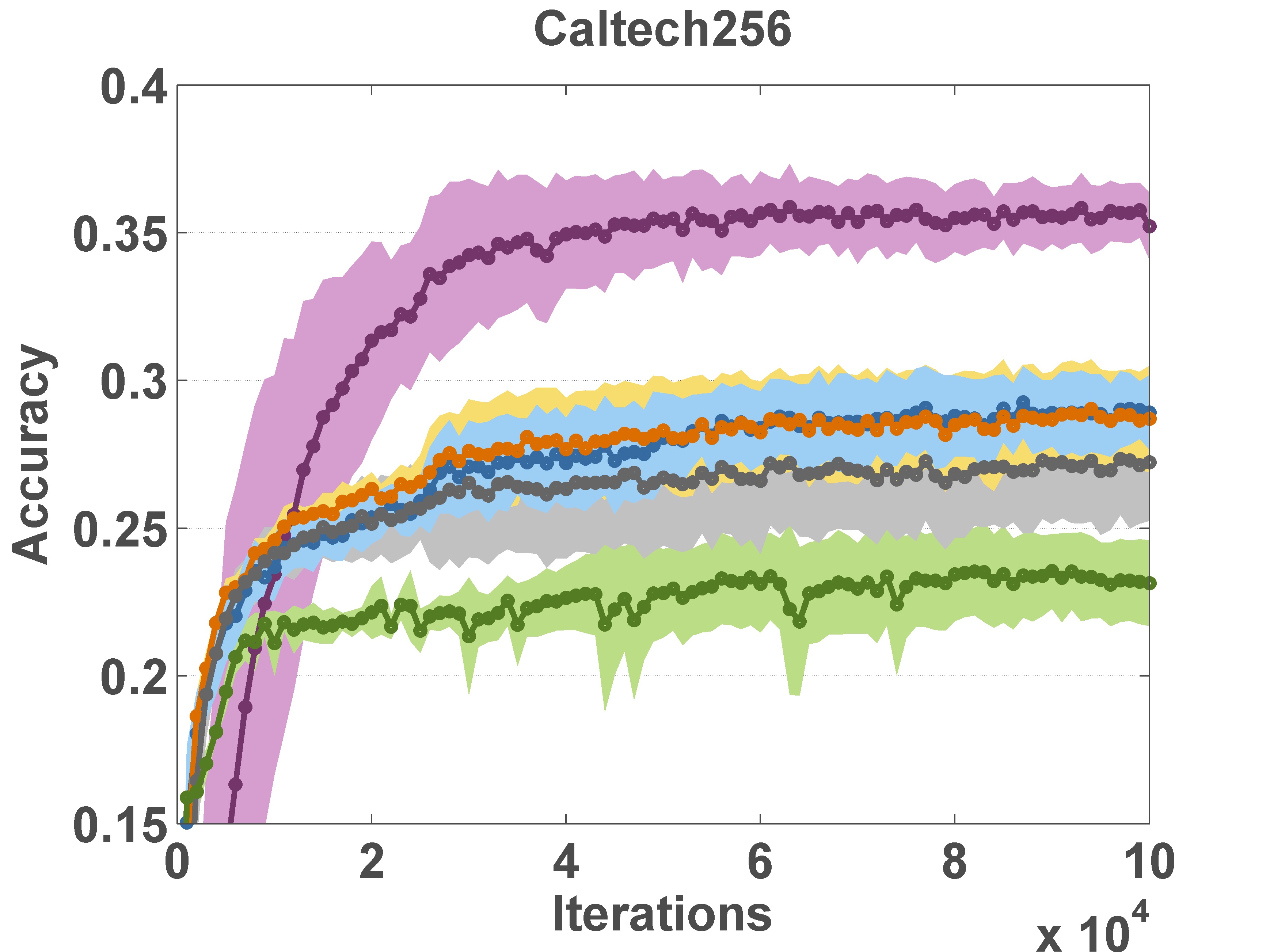}
        \includegraphics[width=0.4\textwidth,trim={0cm 0cm 0.5cm 0cm},clip]{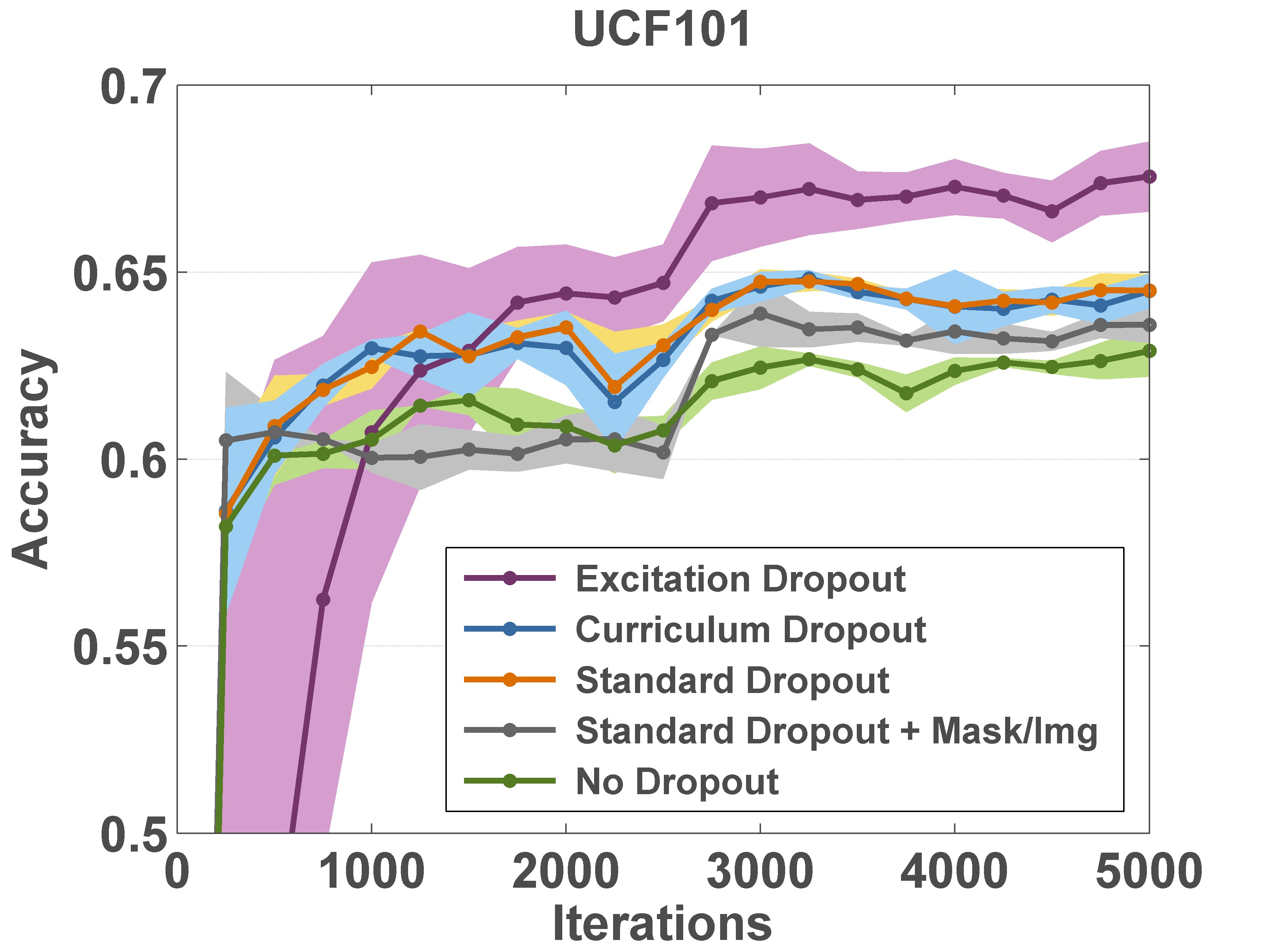}
    
    \caption{We compare the test accuracy of different dropout training strategies on four image/video recognition datasets: Cifar10, Cifar100, Caltech256, UCF101. Results presented here are averaged over five trained models and the standard deviation is depicted around the mean curve using a lighter shade.
    Excitation Dropout is a data-dependent approach. At early iterations, the model is at early stages of optimizing its parameters, and hence more uniform/random saliency in the first iterations. This means that we are not as aggressively regularizing at first, and hence Excitation Dropout takes longer before its effect kicks in, after which it out-performs other techniques.}
    \label{fig:from_scratch}
\end{figure*}

\begin{table*}[t]
\centering
\begin{tabular}{l|c|c|c|c|c} 
\hline
\textbf{Architecture} & \makecell{\textbf{No} \\ \textbf{Dropout (\%)}} & \makecell{\textbf{Standard} \\ \textbf{Dropout \scriptsize{\textit{first fc} (\%)}}} & \makecell{\textbf{Standard} \\ \textbf{Dropout \scriptsize{\textit{both fcs} (\%)}}} & \makecell{\textbf{Curriculum} \\ \textbf{Dropout (\%)}} & \makecell{\textbf{Excitation} \\ \textbf{Dropout (\%)} }\\ 
\hline
\textbf{VGG16} & 69.37 & 71.93 (+2.56\%) & 73.01 (+ 3.64\%) & 72.14 (+2.77\%) & \textbf{73.23 (+3.86\%)} \\
\textbf{VGG19} & 71.32 & 72.52 (+1.29\%) & 73.40 (+ 2.08\%) & 73.18 (+1.86\%) & \textbf{74.34 (+3.02\%)} \\
\textbf{AlexNet} & 62.89 & 64.50 (+1.61\%) & 65.83 (+ 2.94\%) & 64.55 (+1.66\%) &\textbf{67.56 (+4.67\%)}\\
\hline
\end{tabular}
\caption{Test accuracy comparison between No, Standard, Curriculum and Excitation Dropout in the three architectures: AlexNet, VGG16 and VGG19, fine-tuned for the action recognition task on UCF101. The numbers reported are the test accuracies together with improvements (in parenthesis) with respect to No Dropout, averaged over five trained models.}
\label{table:fine_tuning}
\end{table*}

Fig.~\ref{fig:prob_retain} shows $p$ as a function of $p_{EB}$, as in Eqn.~\ref{eq:EBdrop} for $N = 10$ and $P=0.5$. Eqn.~\ref{eq:EBdrop} was designed to fit the following three constraints starting from a general hyperbolic function. Specifically, 1) if neuron $a_i$ has $p_{EB}(a_i)=1$, this results in a retaining probability of $p=0$. We do not want to keep a neuron that has a high contribution to the correct label. 2) If neuron $a_i$ has $p_{EB}(a_i)=0$, this results in a retaining probability of $p=1$. We want to keep a neuron that has not contributed to the correct classification of an image. 3) If neuron $a_i$ has $p_{EB}(a_i)=1/N$, \ie~$p_{EB}$ is a uniform probability distribution, this results in a retaining probability $p=P$. We want to keep a neuron with \textit{base} probability $P$ since all neurons contribute equally. A different choice of the monotonic function that satisfies the three constraints does not change the general pipeline.

Eqn.~\ref{eq:EBdrop} provides a dropout probability for each neuron, which is then used as the parameter of a Bernoulli distribution giving a binary dropout mask. During training, each image in a batch leads to different excitatory connections in the network and therefore has a different $p_{EB}$ distribution, consequently leading to a different dropout mask. Following the standard practical implementations for Standard Dropout, when Excitation Dropout is employed during training, the activations are rescaled  according to the fraction of dropped neurons, and no dropout or rescaling is employed at test time. Fig.~\ref{fig:pipeline} presents the pipeline of Excitation Dropout at training time, and a run-time analysis is presented in Section~\ref{act}.


\section{Experiments}
\label{exps}
In this section, we present how Excitation Dropout improves the generalization ability on four image/video recognition datasets in fully connected layers of different architectures. We then present an analysis of how Excitation Dropout affects the utilization of network neurons on the same datasets. Finally, we examine the resilience of a model trained using Excitation Dropout to network compression.

\subsection{Datasets and Architectures}
We present results on four image/video recognition datasets. \textbf{Cifar10} and \textbf{Cifar100} \cite{krizhevsky2009learning} are image recognition datasets, each consisting of $60000$ $32 \times 32$ tiny RGB natural images. Cifar10 images are distributed over $10$ classes with $6000$ images per class, and Cifar100 images are distributed over $100$ classes with $600$ images per class. Standard training and testing splits contain $50K$ and $10K$ images, respectively. We feed the network with the original image dimensions shuffling the training images within the standard defined splits when training multiple models. \textbf{Caltech256} \cite{griffinHolubPerona} is an image recognition dataset consisting $31000$ RGB images divided in $256$ classes. We consider five different random splits of $50$ training images and $20$ testing images for each class. Images were reshaped to $128 \times 128$ pixel to feed the network.
\textbf{UCF101} \cite{soomro2012ucf101} is a video action recognition dataset consisting of 13320 action videos belonging to 101 action classes, from which we sample two million frames. For this dataset we consider a frame-based action recognition task. 
The images are resized to $224\times224$ and $227\times227$ to fit the input layers of the VGG and AlexNet architectures, respectively.

We present results on four architectures. Relatively shallow architectures are trained from scratch, and deeper popular architectures are fine-tuned after being pre-trained on ImageNet \cite{deng2009imagenet}.

\textbf{Models trained from scratch:} We train the CNN-2 architecture used in \cite{morerio2017curriculum}, which adopts the state-of-the-art dropout variant, for comparison purposes. This architecture consists of three convolutional and two fully-connected layers. Table \ref{tab:CNN2} shows the details of the CNN-2 architecture adopted in the experiments. The size of the softmax layer depends upon the number of classes for each dataset. We train this network from scratch for $100K$ iterations using the Adam solver on the datasets: Cifar-10, Cifar-100 and Caltech-256. We use a batch size of 100 images and fix the learning rate to be $10^{-3}$, decreasing to $10^{-4}$ after $25K$ iterations, and a weight decay of 0.005. 

\textbf{Fine-tuned models:} We fine-tune the commonly used architectures: AlexNet \cite{krizhevsky2012imagenet}, VGG16 and VGG19 \cite{simonyan2014very} pre-trained on ImageNet. We fine-tune the models for a frame by frame action recognition task on UCF101 using the Adam solver. The learning rate is fixed to $10^{-3}$ for all the processes. We fine-tune AlexNet for $5K$ while VGG16 and VGG19 for $30K$ iterations. We use a batch size of 128 and 50 images for AlexNet and VGG16/19, respectively, and a weight decay of 0.005.

\begin{table}[t]
\centering
\begin{tabular}{B|D|E|E|E} 
\hline
\multirow{2}{*} & \multirow{2}{*} & \multicolumn{3}{c}{\makecell{\textbf{Acc.(\%)} \\ \textbf{@ Dropout Rate}}} \\
\cline{3-5}
\makecell{\textbf{Dataset} \\ \scriptsize{(Architecture)}} & \makecell{\textbf{Dropout} \\ \textbf{Scheme}} & \makecell{\textbf{0.25}} & \makecell{\textbf{0.5}} & 
\makecell{\textbf{0.75}}\\ 
\hline
\multirow{3}{*}{\makecell{\textbf{\textit{Cifar10}} \\ \scriptsize{(CNN-2)}}} & Standard \scriptsize{\textit{first fc}} & 79.16 &80.13 & 81.19 \\
& Standard \scriptsize{\textit{both fcs}} & 79.47 & 80.80 & \bf{81.72} \\
& Excitation& \bf{81.38} & \bf{81.94} & 81.55 \\
\hline
\multirow{3}{*}{\makecell{\textbf{\textit{Cifar100}} \\ \scriptsize{(CNN-2)}}} & Standard \scriptsize{\textit{first fc}} & 48.44 &50.36 & 51.64 \\
& Standard \scriptsize{\textit{both fcs}} & 50.30 & 51.96 & 49.46 \\
& Excitation & \bf{53.23} & \bf{52.04} & \bf{51.87}  \\
\hline
\multirow{3}{*}{\makecell{\textbf{\textit{Caltech256}} \\ \scriptsize(CNN-2)}} & Standard \scriptsize{\textit{first fc}} & 26.23 &28.73 & 32.51 \\
& Standard \scriptsize{\textit{both fcs}} & 27.70 & 31.77 & 33.54 \\
& Excitation & \bf{33.60} & \bf{35.77} &\bf{36.81} \\
\hline
\multirow{3}{*}{\makecell{\textbf{\textit{UCF101}} \\ \scriptsize{(VGG16)}}} & Standard \scriptsize{\textit{first fc}} & 71.01 &71.93 & 72.92 \\
& Standard \scriptsize{\textit{both fcs}} & 71.48 & 73.01 & 72.89 \\
& Excitation & \bf{73.56} & \bf{73.23} &\bf{73.06} \\
\hline
\end{tabular}
\caption{Hyper-parameter sensitivity analysis for the Standard Dropout probability, and the Excitation Dropout base dropout probability. The accuracy is reported on the test set of each dataset. The retaining probability $p$ or $P$ is one minus the dropout rate.}
\label{table:sensitivity}
\end{table}

\begin{figure*}[h!]
	\centering
	Neurons ON \hspace*{12em} Peak $p_{EB}$\\[0.2em]
	\includegraphics[width=0.33\textwidth,trim={0.1cm 0cm 0.7cm 0cm},clip]{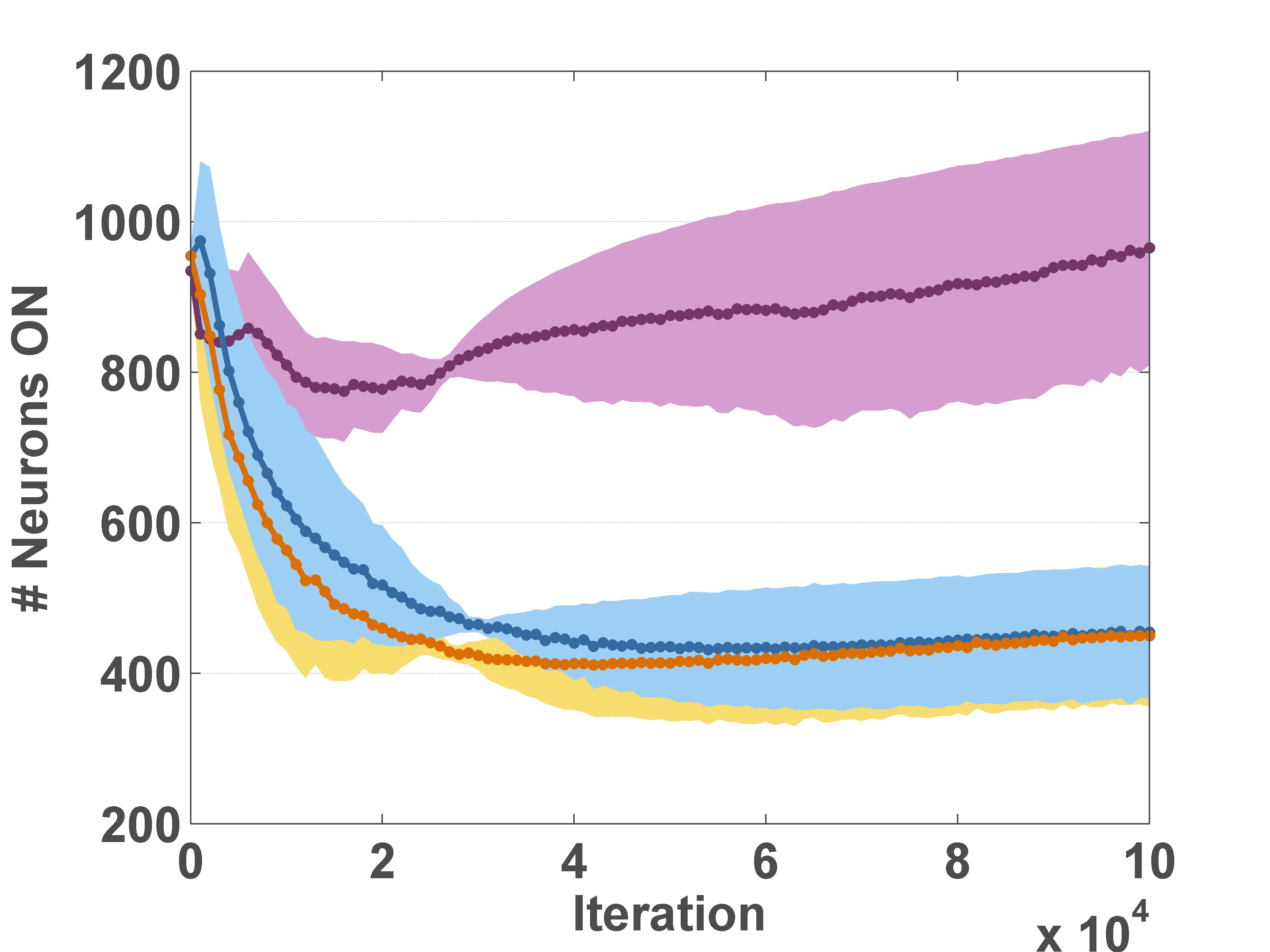}
	\includegraphics[width=0.33\textwidth,trim={0.2cm 0cm 0.7cm 0cm},clip]{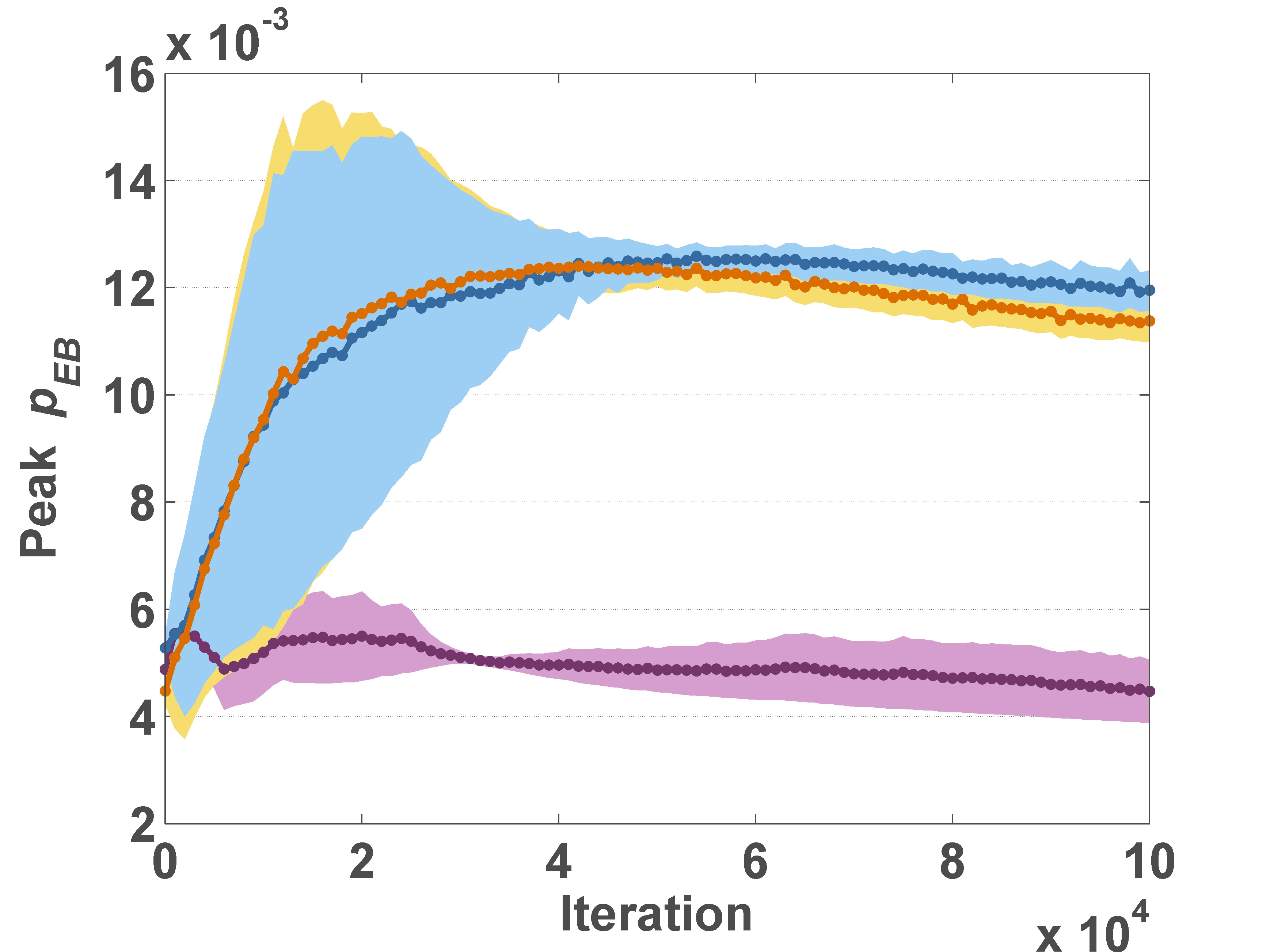}\\[1.2em]
	Entropy of Activations \hspace*{8em} Entropy of $p_{EB}$\\[0.2em]
	\includegraphics[width=0.33\textwidth,trim={0.1cm 0cm 0.6cm 0.5cm},clip]{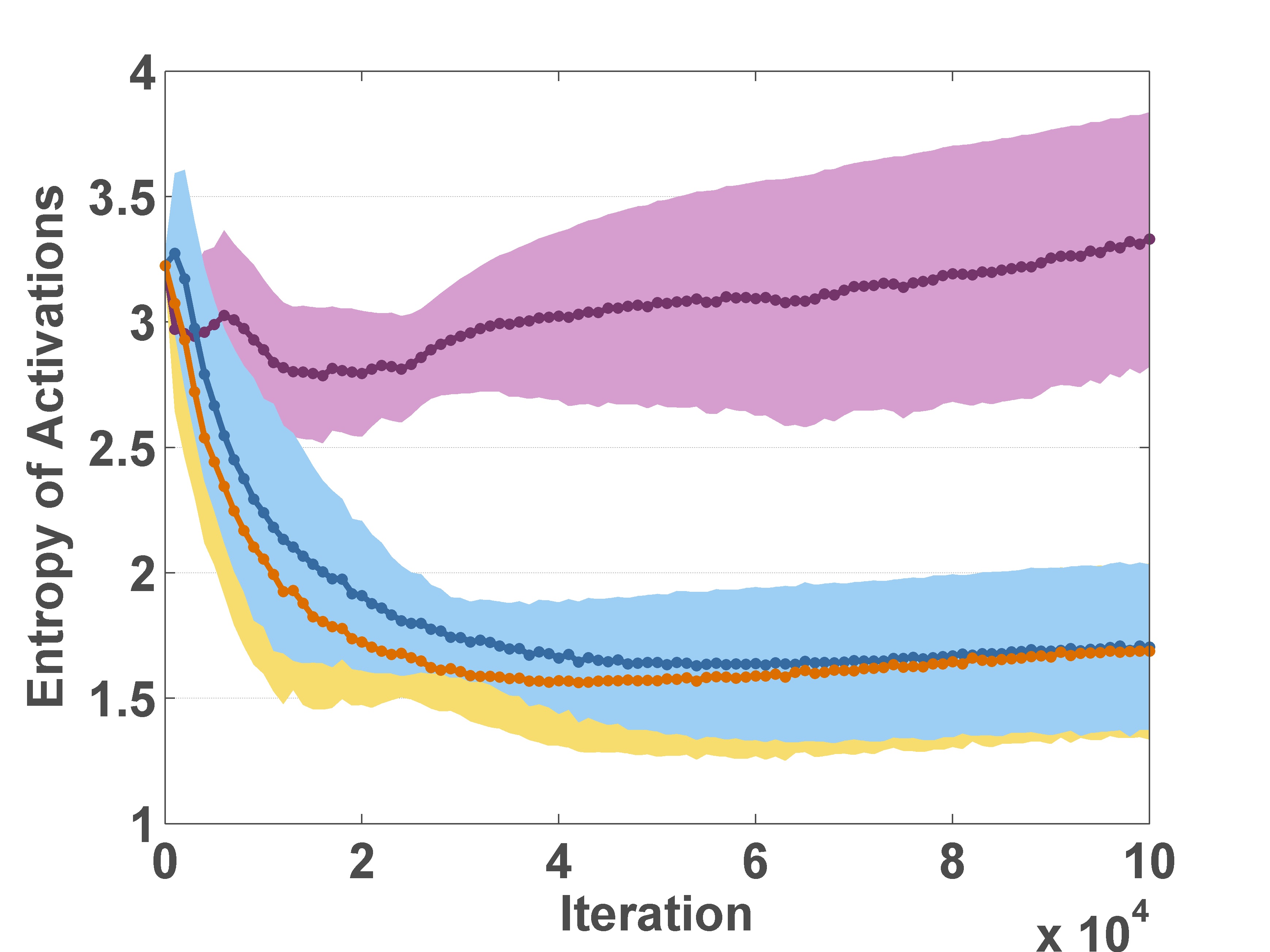}
	\includegraphics[width=0.33\textwidth,trim={0.2cm 0cm 0.6cm 0.5cm},clip]{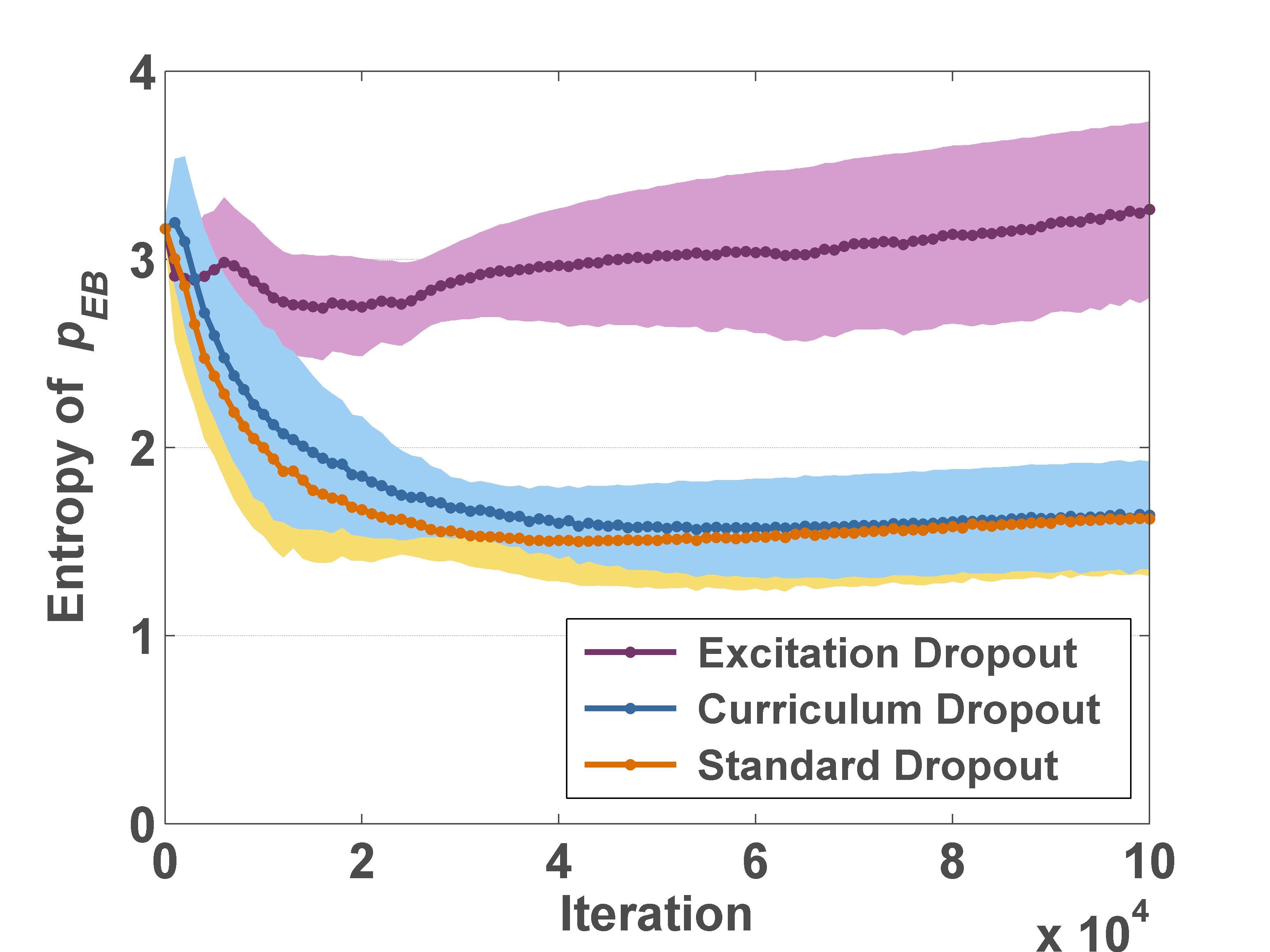} \vspace{0.5 em}
	\caption{\textbf{Cifar100.} \textit{\# Neurons ON}, \textit{Peak $p_{EB}$}, \textit{Entropy of Activations}, and \textit{Entropy of $p_{EB}$} over time during training.}
	\label{fig:metrics_over_time_Cifar100}
\end{figure*}


\begin{figure*}[h!]
	\centering
	Neurons ON \hspace*{12em} Peak $p_{EB}$\\[0.2em]
	\includegraphics[width=0.33\textwidth,trim={0cm 0cm 0.7cm 0cm},clip]{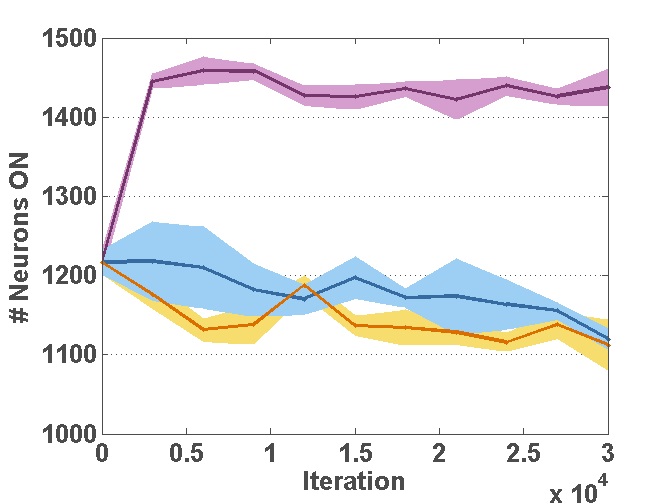}
	\includegraphics[width=0.33\textwidth,trim={0cm 0cm 0.7cm 0cm},clip]{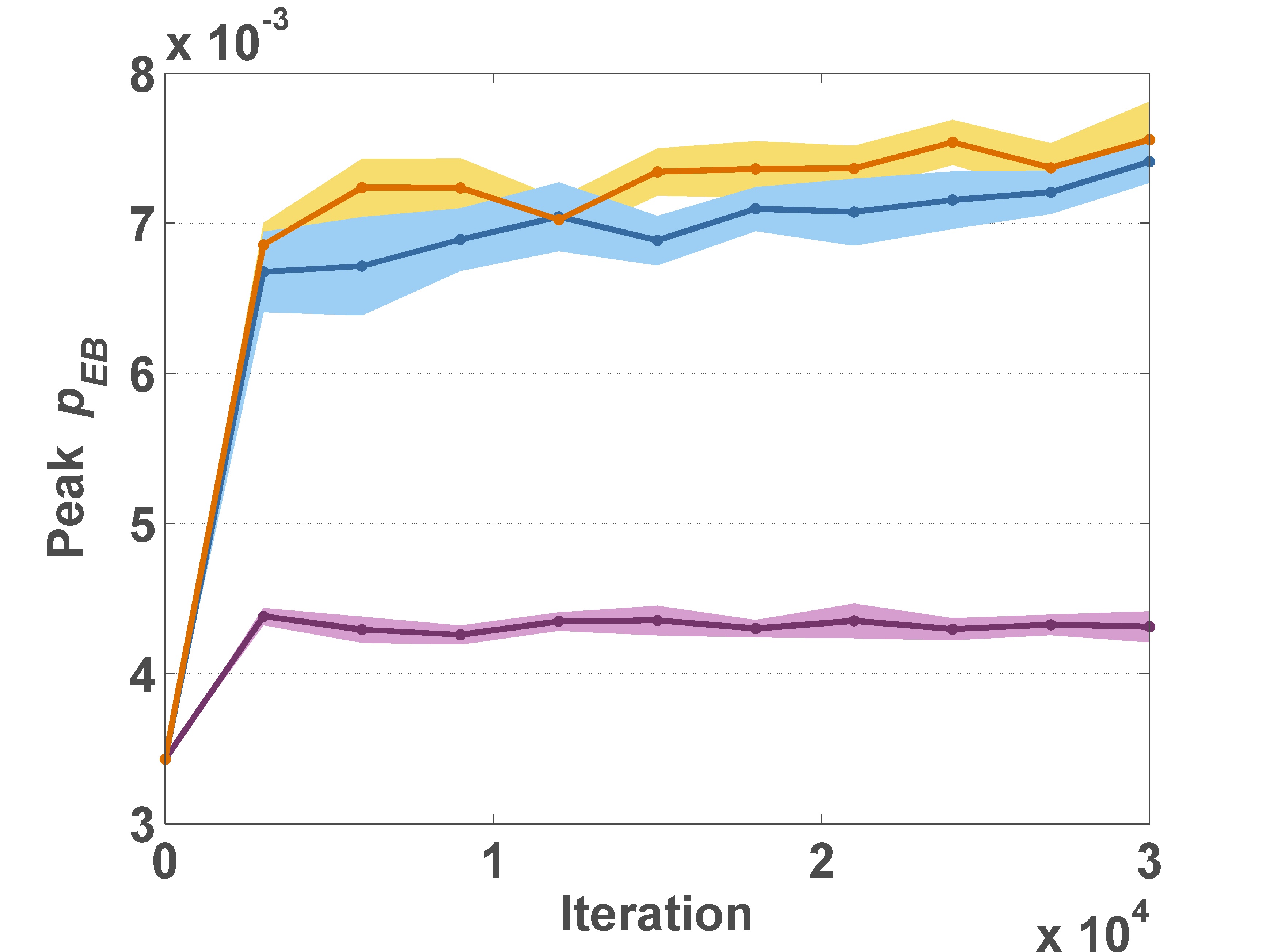}\\[0.5em]
	Entropy of Activations \hspace*{8em} Entropy of $p_{EB}$\\[1.2em]
	\includegraphics[width=0.33\textwidth,trim={0cm 0cm 0.6cm 0.5cm},clip]{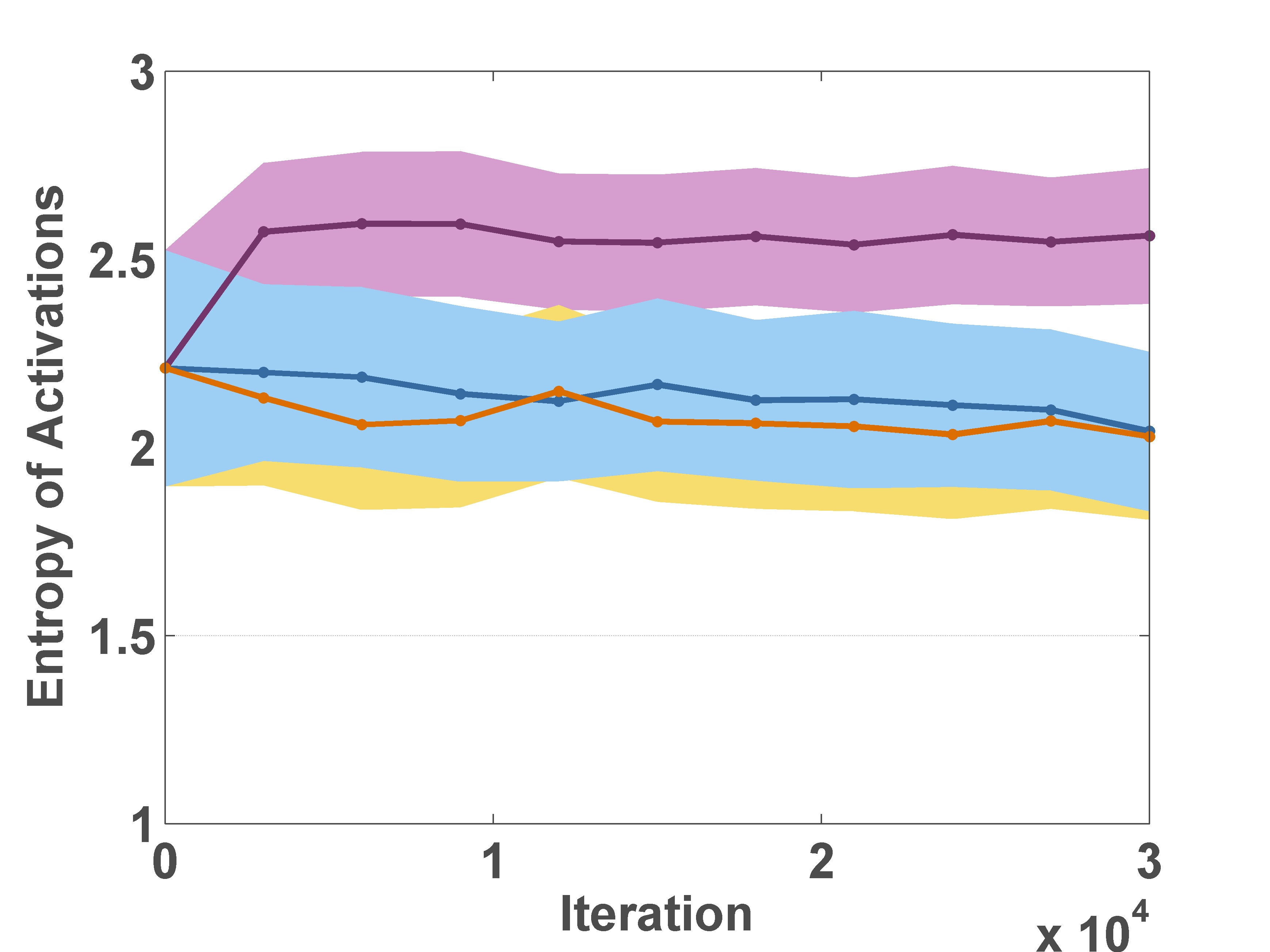}
	\includegraphics[width=0.33\textwidth,trim={0cm 0cm 0.6cm 0.5cm},clip]{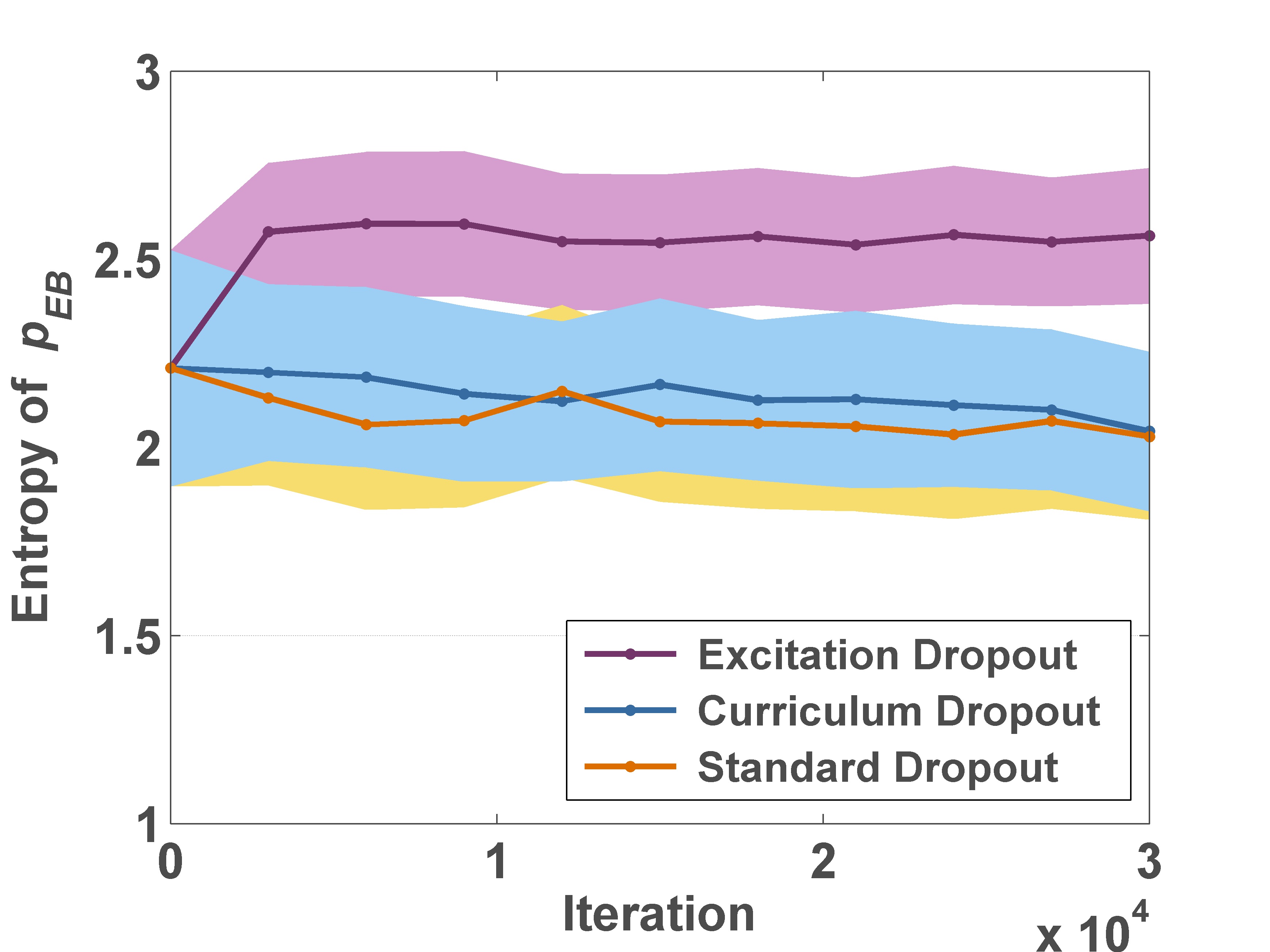}\vspace{0.5 em}
	\caption{\textbf{UCF101.} \textit{\# Neurons ON}, \textit{Peak $p_{EB}$}, \textit{Entropy of Activations}, and \textit{Entropy of $p_{EB}$} over time during training.}
	\label{fig:metrics_over_time_UCF}
\end{figure*}


\begin{table*}[t]
\centering
\begin{tabular}{c|c|c|c|c} 
\hline
\makecell{\textbf{Dataset} \\ \scriptsize{(Architecture)}} &  \textbf{Metric} & \makecell{\textbf{Standard} \\ \textbf{Dropout}} & \makecell{\textbf{Curriculum} \\ \textbf{Dropout}}& \makecell{\textbf{Excitation} \\ \textbf{Dropout}} \\
\hline
\multirow{2}{*}{\rotatebox[origin=c]{90}{\makecell{\textbf{\textit{Cifar10}} \\ \scriptsize{(CNN-2)}}\hspace*{0.9em}}}  & \# Neurons ON & 1194 {\small ($\pm 153$)} & 1169 {\small ($\pm 61$)} & \textbf{1325} {\small ($\pm 61$)} \\ 
& Peak $p_{EB}$ & 0.011 {\small ($\pm 0.004$)} & 0.009 {\small ($\pm 0.001$)} & \textbf{0.003} {\small ($\pm 0.0002$)} \\ 
& Entropy of Activations & 3.55 {\small ($\pm 0.72$)} & 3.50 {\small ($\pm 0.12$)} &  \textbf{4.29} {\small ($\pm 0.28$)} \\ 
& Entropy of $p_{EB}$ & 3.28 {\small ($\pm 0.56$)} & 3.32 {\small ($\pm 0.13$)} &\textbf{4.26} {\small ($\pm 0.26$)} \\ 
& Conservative Filters$_{\Delta=0.25}$ & 1204 {\small ($\pm 37$)} & 959 {\small ($\pm 34 $)} & \textbf{124} {\small ($\pm 22$)} \\ 
\hline
\multirow{2}{*}{\rotatebox[origin=c]{90}{\makecell{\textbf{\textit{Cifar100}} \\ \scriptsize{(CNN-2)}}\hspace*{0.6em}}} & \# Neurons ON & 453 {\small ($\pm 183$)} & 460 {\small ($\pm 75$)} & \textbf{943} {\small ($\pm 131$)} \\ 
& Peak $p_{EB}$ & 0.011 {\small ($\pm 0.0004$)} & 0.012 {\small ($\pm 0.0004$)} &\textbf{0.005} {\small ($\pm 0.0005$)} \\ 
& Entropy of Activations & 1.67 {\small ($\pm 0.31$)} & 1.70 {\small ($\pm 0.29$)} &\textbf{3.21} {\small ($\pm 0.44$)}  \\ 
& Entropy of $p_{EB}$ & 1.64 {\small ($\pm 0.27$)} & 1.67 {\small ($\pm 0.26$)} & \textbf{3.17} {\small ($\pm 0.41$)} \\ 
& Conservative Filters$_{\Delta=0.30}$ & 2048 {\small ($\pm 51 $)} & 2038 ($\pm 44 $) & \textbf{14} {\small ($\pm 13 $)}\\ 
\hline
\multirow{2}{*}{\rotatebox[origin=c]{90}{\makecell{\textbf{\textit{Caltech256}} \\ \scriptsize{(CNN-2)}}\hspace*{0.2em}}} & \# Neurons ON & 412 {\small ($\pm 126$)} & 471 {\small ($\pm 146$)} & \textbf{702} {\small ($\pm 171$)} \\ 
& Peak $p_{EB}$ & 0.014 {\small ($\pm 0.0007$)} & 0.013 {\small ($\pm 0.0006$)} & \textbf{0.007} {\small ($\pm 0.0003$)} \\ 
& Entropy of Activations & 1.63 {\small ($\pm 0.32 $)} & 1.84 {\small ($\pm 0.35$)} & \textbf{2.63} {\small ($\pm 0.23$)} \\ 
& Entropy of $p_{EB}$ & 1.58 {\small ($\pm 0.29$)} & 1.77 {\small ($\pm 0.31$)} & \textbf{2.59} {\small ($\pm 0.22$)} \\ 
& Conservative Filters$_{\Delta=1.25}$ & 2048 ($\pm 46 $) & 2048 ($\pm 49 $)& \textbf{1671} ($\pm 31 $)\\ 
\hline
\multirow{2}{*}{\rotatebox[origin=c]{90}{\makecell{\textbf{\textit{UCF101}} \\ \scriptsize{(VGG16)}}\hspace*{0.7em}}} & \# Neurons ON & 1120 {\small ($\pm 25$)} & 1143 {\small ($\pm 22$)} & \textbf{1404} {\small ($\pm 37$)} \\ 
& Peak $p_{EB}$ & 0.007 {\small ($\pm 0.0002$)} & 0.007 {\small ($\pm 0.0002$)} &\textbf{0.004} {\small ($\pm 0.0002$)} \\ 
& Entropy of Activations & 2.04 {\small ($\pm 0.23$)}  & 2.08 {\small ($\pm 0.21$)} &\textbf{2.51} {\small ($\pm 0.18$)} \\ 
& Entropy of $p_{EB}$ & 1.92 {\small ($\pm 0.22$)} & 1.95 {\small ($\pm 0.20$)} &\textbf{2.42} {\small ($\pm 0.18$)} \\ 
& Conservative Filters$_{\Delta=0.15}$ & 3599 ($\pm 66 $) &3859 ($\pm 53 $) & \textbf{44} ($\pm 36 $) \\ 
\hline
\end{tabular}
\caption{Different metrics to reflect the usage of network capacity in the first fully-connected layer of the CNN-2 architecture consisting of 2048 neurons and the VGG16 consisting of 4096 neurons. Results presented here are averaged over five trained models for each of the datasets: Cifar10, Cifar100, Caltech256 and UCF101 ($\sigma$ in brackets). Excitation Dropout consistently produces more neurons with non-zero activations, has a more spread saliency map leading to a lower saliency peak, has a higher entropy of both activations and saliency, and has a lower number of conservative filters; all reflecting an improved utilization of the network neurons using Excitation Dropout.}
\vspace*{-1em}
\label{table:metrics}
\end{table*}


\begin{figure*}[t!]
\centering
\vspace*{1.4em}
\textbf{Excitation Dropout}  \\
\includegraphics[width=0.115\linewidth,height=0.115\linewidth,trim={0.5cm 0.4cm 1.4cm 1cm},clip]{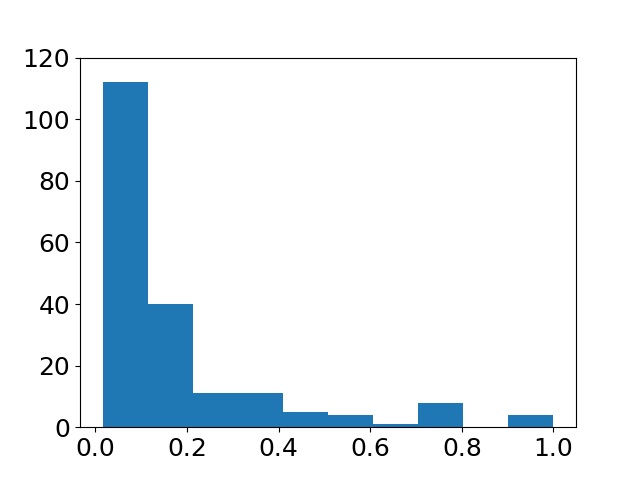}
\hspace*{0.5em}
\includegraphics[width=0.115\linewidth,height=0.115\linewidth,trim={3cm 0cm 3cm 0cm},clip]{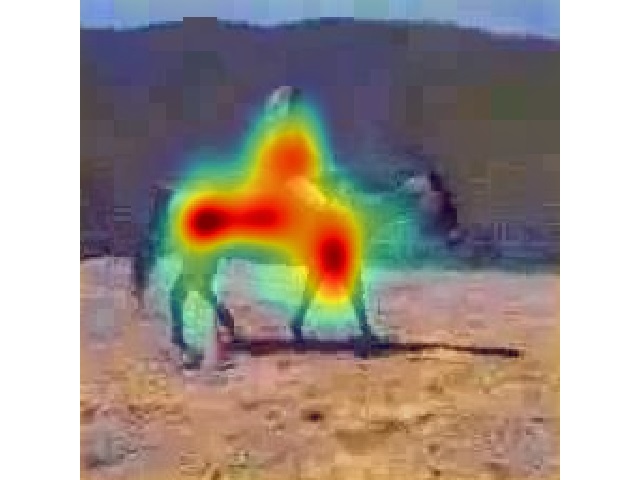}
\includegraphics[width=0.115\linewidth,height=0.115\linewidth,trim={3cm 0cm 3cm 0cm},clip]{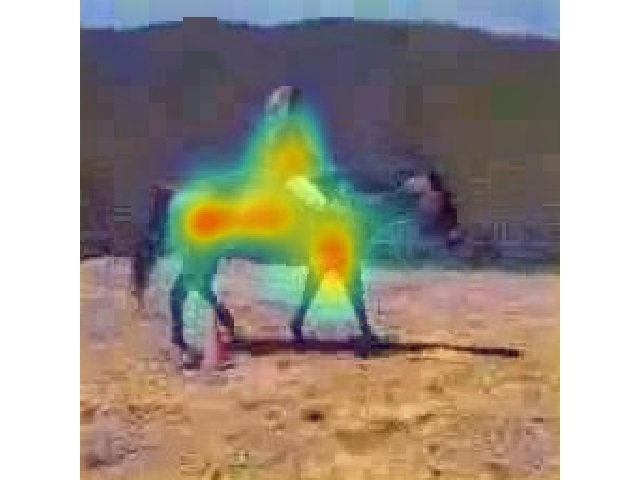}
\includegraphics[width=0.115\linewidth,height=0.115\linewidth,trim={3cm 0cm 3cm 0cm},clip]{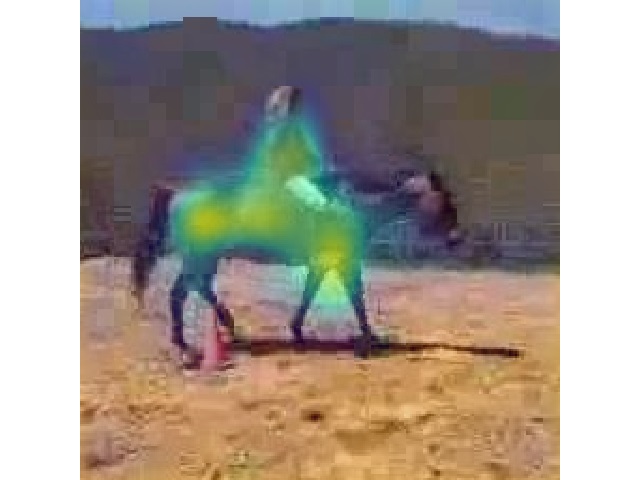}
\includegraphics[width=0.115\linewidth,height=0.115\linewidth,trim={3cm 0cm 3cm 0cm},clip]{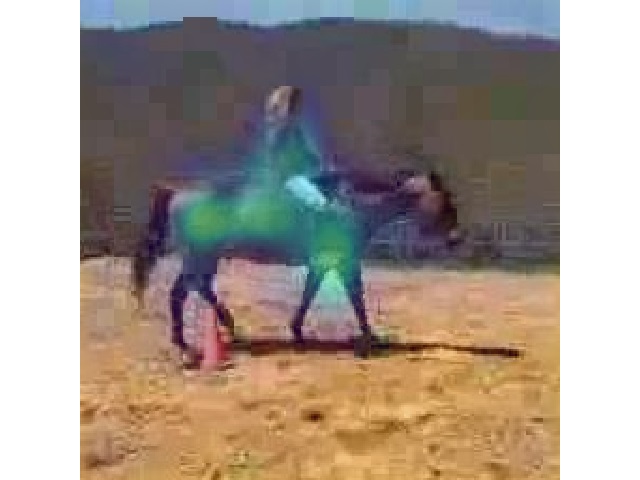}
\includegraphics[width=0.115\linewidth,height=0.115\linewidth,trim={3cm 0cm 3cm 0cm},clip]{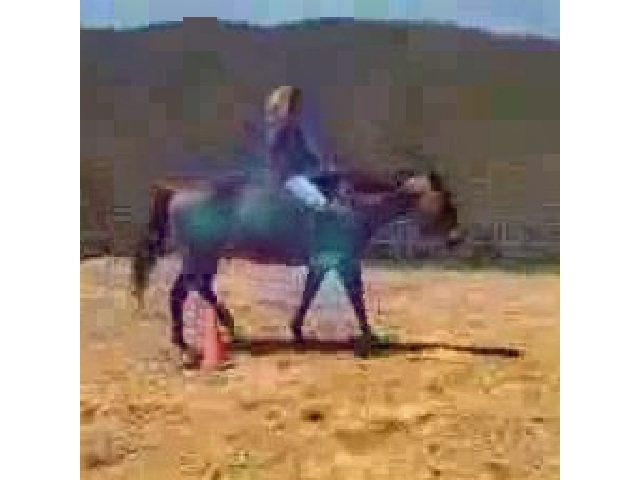}
\includegraphics[width=0.115\linewidth,height=0.115\linewidth,trim={3cm 0cm 3cm 0cm},clip]{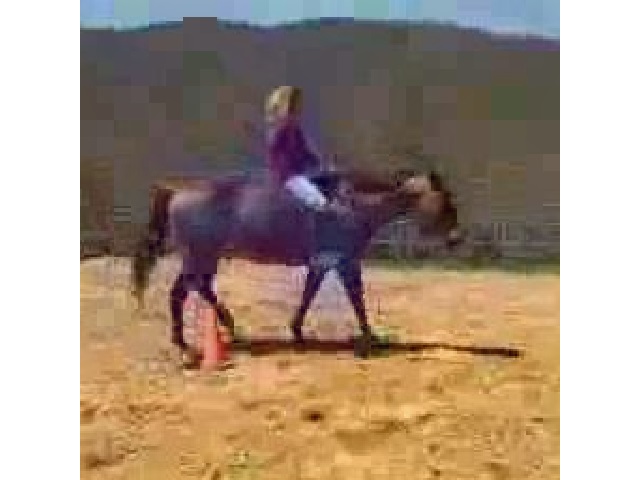}
\hspace*{0.5em}
\includegraphics[width=0.115\linewidth,height=0.115\linewidth,trim={0.5cm 0.4cm 1.4cm 1cm},clip]{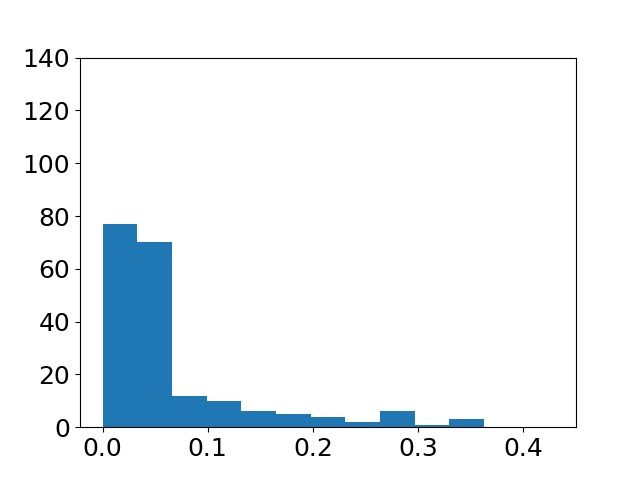}\\[0.5em]
\textbf{Curriculum Dropout}  \\
\includegraphics[width=0.115\linewidth,height=0.115\linewidth,trim={0.5cm 0.4cm 1.4cm 1cm},clip]{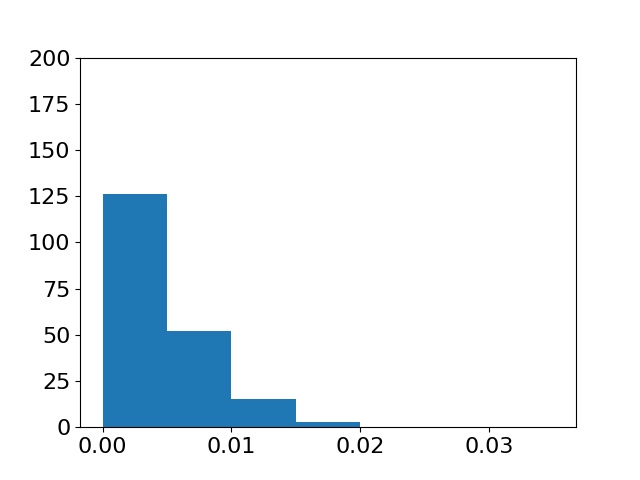}
\hspace*{0.5em}
\includegraphics[width=0.115\linewidth,height=0.115\linewidth,trim={3cm 0cm 3cm 0cm},clip]{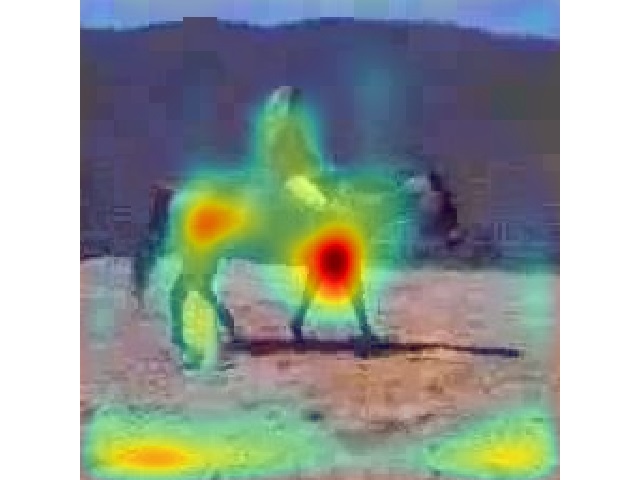}
\includegraphics[width=0.115\linewidth,height=0.115\linewidth,trim={3cm 0cm 3cm 0cm},clip]{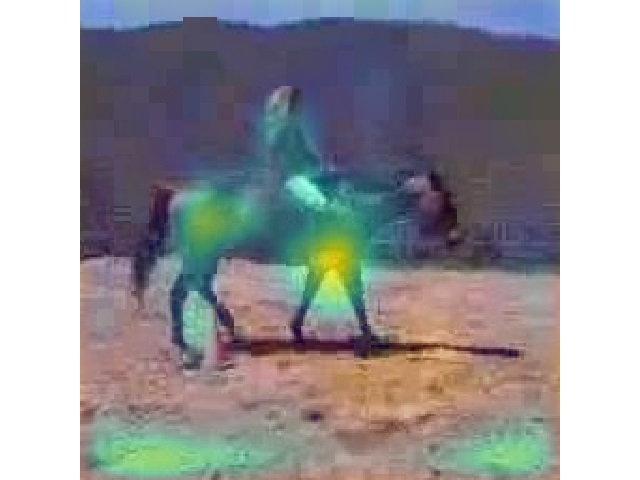}
\includegraphics[width=0.115\linewidth,height=0.115\linewidth,trim={3cm 0cm 3cm 0cm},clip]{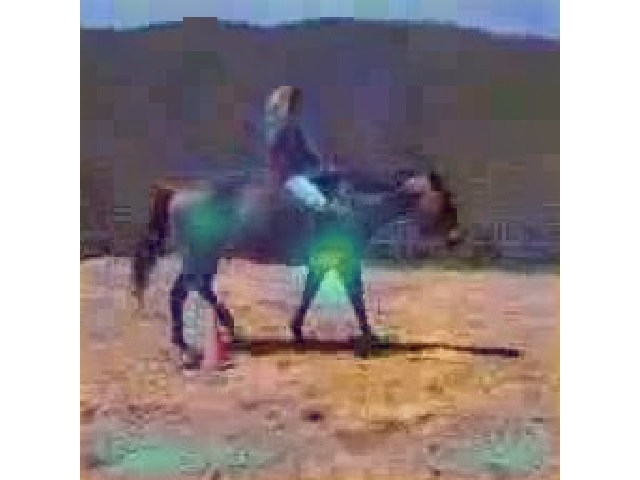}
\includegraphics[width=0.115\linewidth,height=0.115\linewidth,trim={3cm 0cm 3cm 0cm},clip]{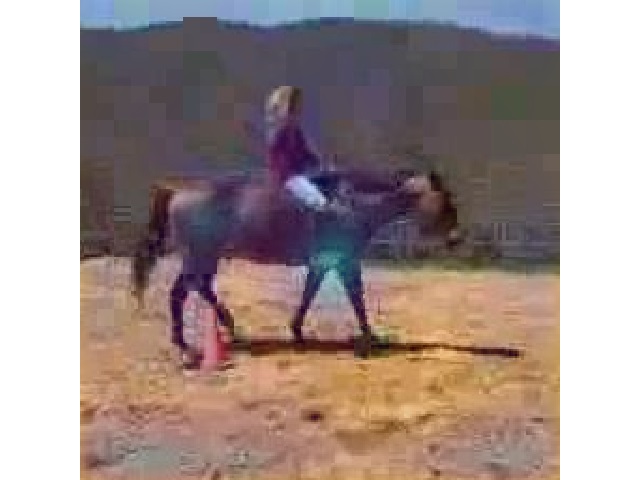}
\includegraphics[width=0.115\linewidth,height=0.115\linewidth,trim={3cm 0cm 3cm 0cm},clip]{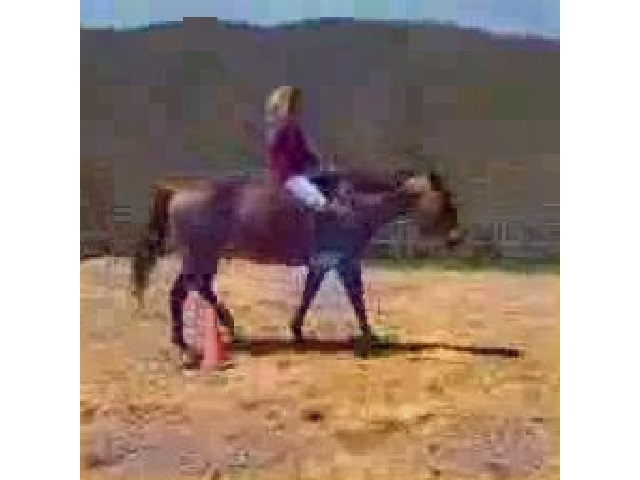}
\includegraphics[width=0.115\linewidth,height=0.115\linewidth,trim={3cm 0cm 3cm 0cm},clip]{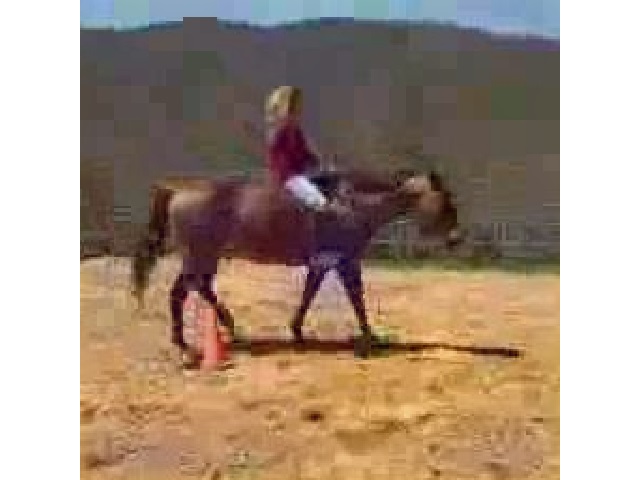} 
\hspace*{0.5em}
\includegraphics[width=0.115\linewidth,height=0.115\linewidth,trim={0.5cm 0.4cm 1.4cm 1cm},clip]{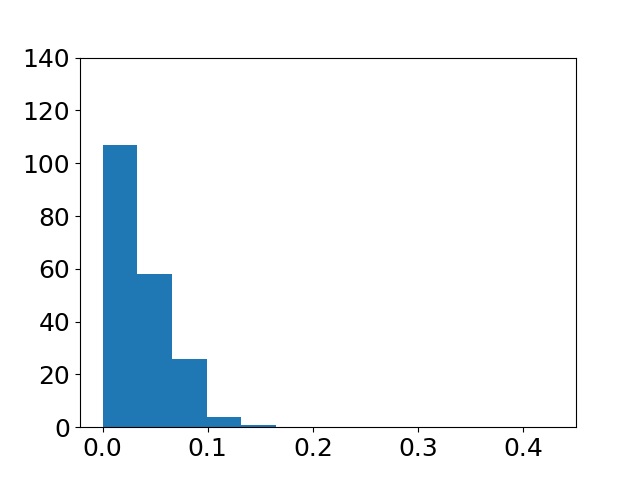}\\[0.5em]
\textbf{Standard Dropout}  \\
\includegraphics[width=0.115\linewidth,height=0.115\linewidth,trim={0.5cm 0.4cm 1.4cm 1cm},clip]{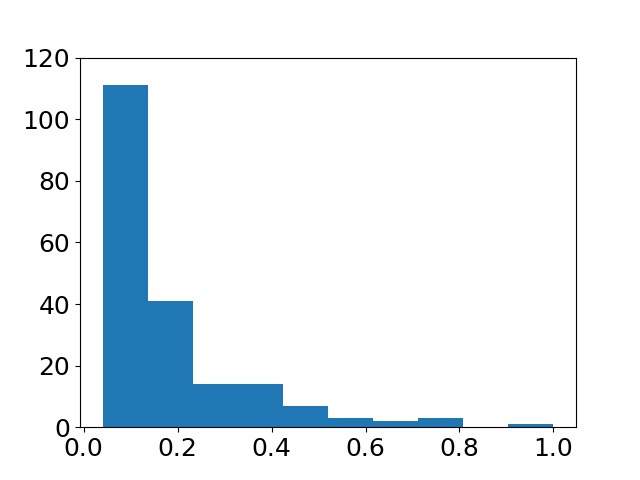}
\hspace*{0.5em}
\includegraphics[width=0.115\linewidth,height=0.115\linewidth,trim={3cm 0cm 3cm 0cm},clip]{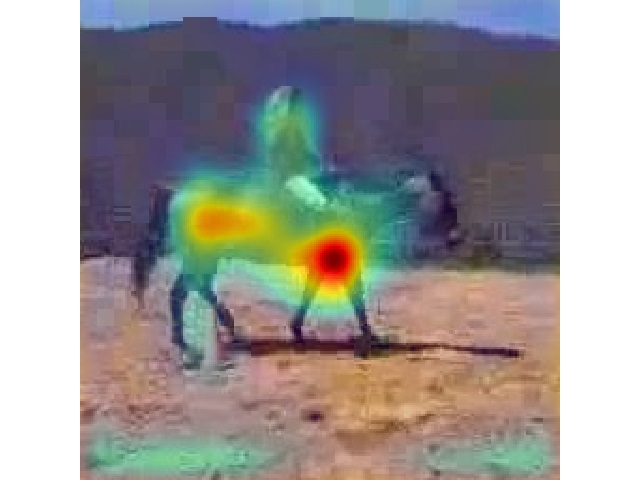}
\includegraphics[width=0.115\linewidth,height=0.115\linewidth,trim={3cm 0cm 3cm 0cm},clip]{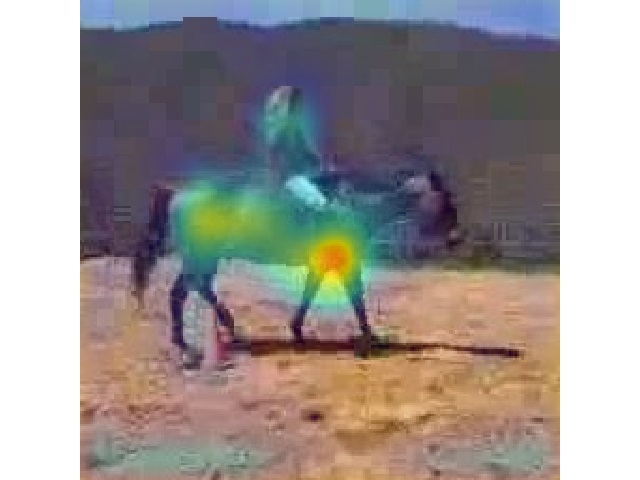}
\includegraphics[width=0.115\linewidth,height=0.115\linewidth,trim={3cm 0cm 3cm 0cm},clip]{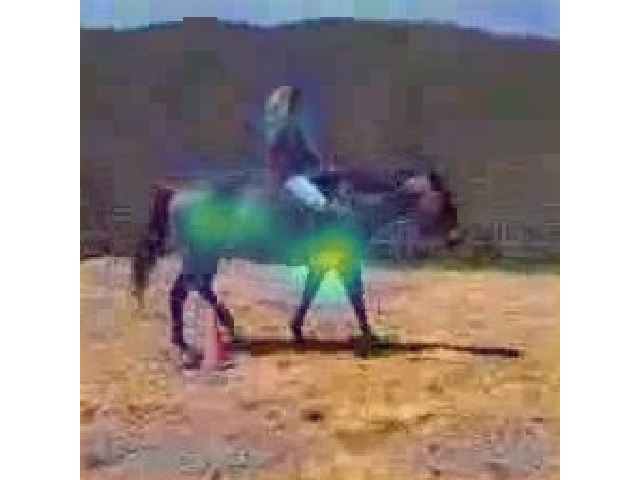}
\includegraphics[width=0.115\linewidth,height=0.115\linewidth,trim={3cm 0cm 3cm 0cm},clip]{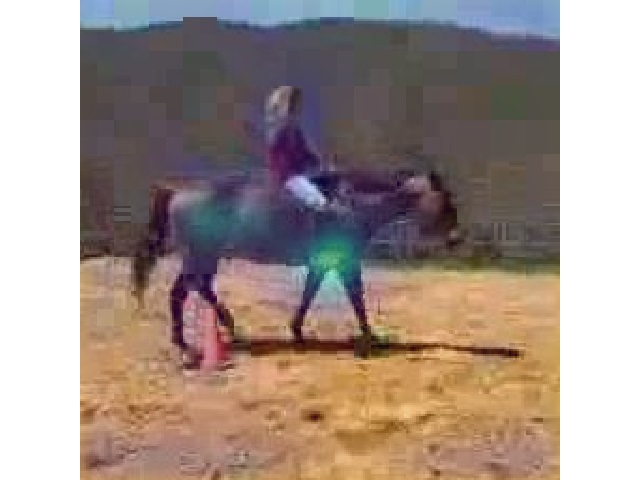}
\includegraphics[width=0.115\linewidth,height=0.115\linewidth,trim={3cm 0cm 3cm 0cm},clip]{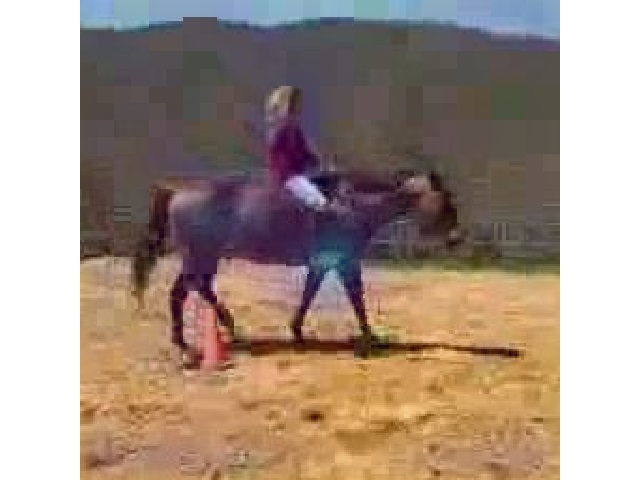}
\includegraphics[width=0.115\linewidth,height=0.115\linewidth,trim={3cm 0cm 3cm 0cm},clip]{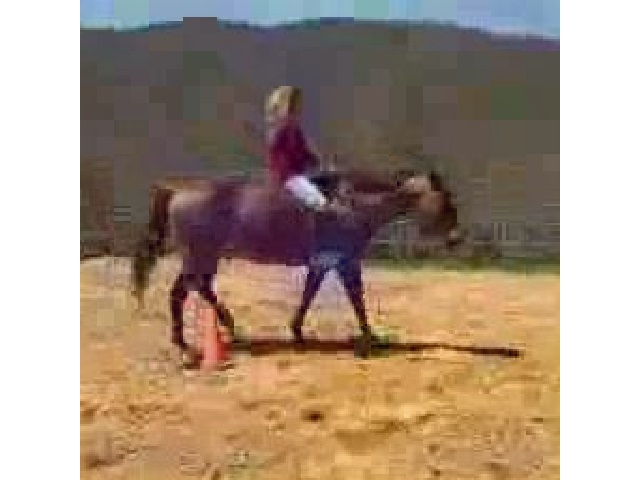}
\hspace*{0.5em}
\includegraphics[width=0.115\linewidth,height=0.115\linewidth,trim={0.5cm 0.4cm 1.4cm 1cm},clip]{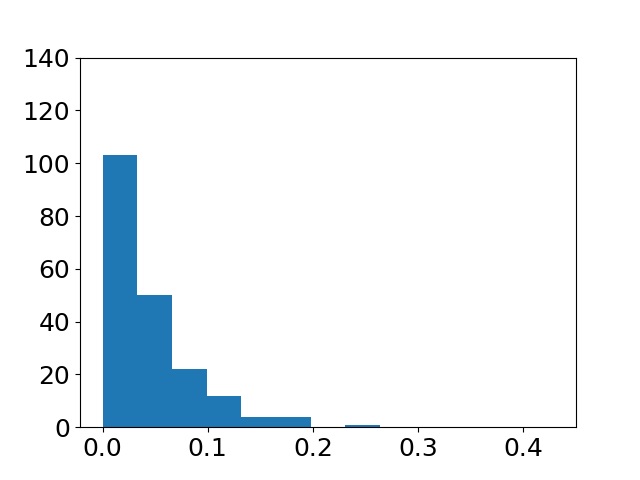}\\[0.5em]
\textbf{No Dropout}  \\
\includegraphics[width=0.115\linewidth,height=0.115\linewidth,trim={0.5cm 0.4cm 1.4cm 1cm},clip]{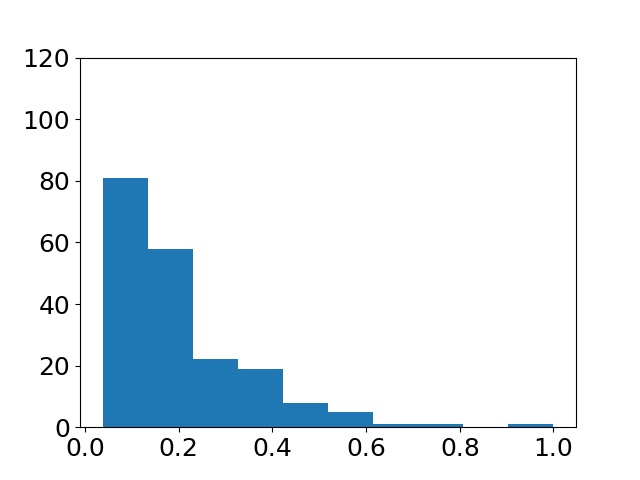}
\hspace*{0.5em}
\includegraphics[width=0.115\linewidth,height=0.115\linewidth,trim={3cm 0cm 3cm 0cm},clip]{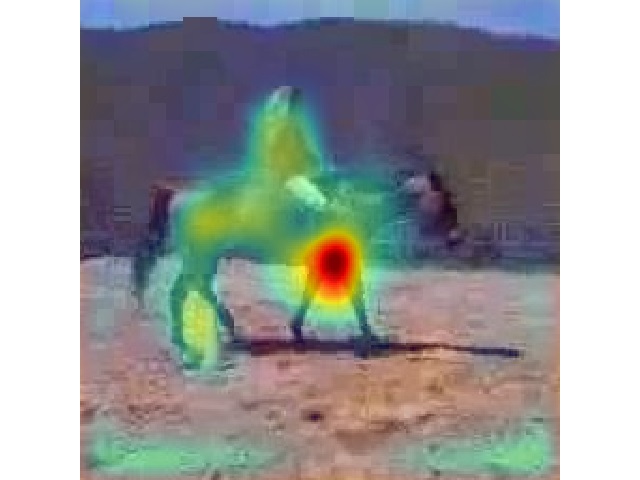}
\includegraphics[width=0.115\linewidth,height=0.115\linewidth,trim={3cm 0cm 3cm 0cm},clip]{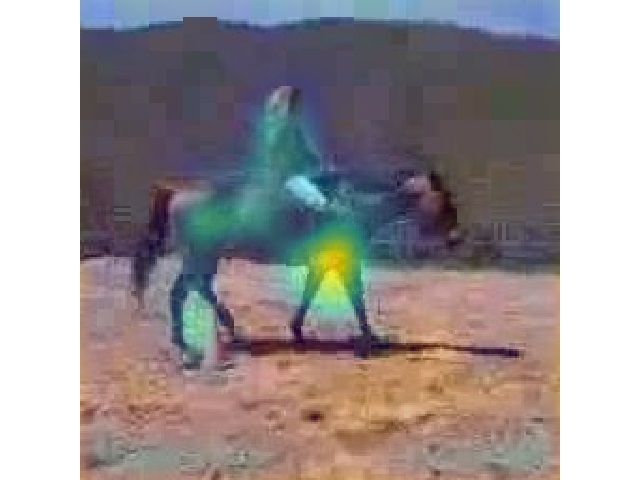}
\includegraphics[width=0.115\linewidth,height=0.115\linewidth,trim={3cm 0cm 3cm 0cm},clip]{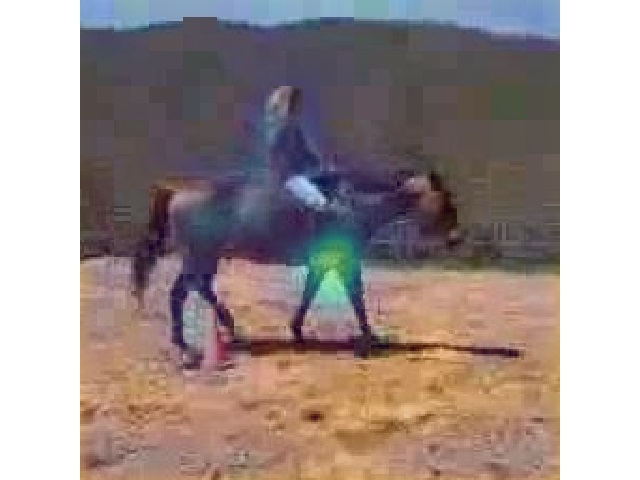}
\includegraphics[width=0.115\linewidth,height=0.115\linewidth,trim={3cm 0cm 3cm 0cm},clip]{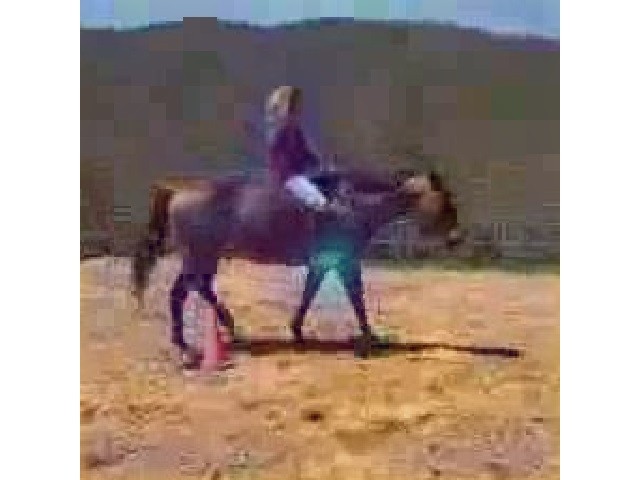}
\includegraphics[width=0.115\linewidth,height=0.115\linewidth,trim={3cm 0cm 3cm 0cm},clip]{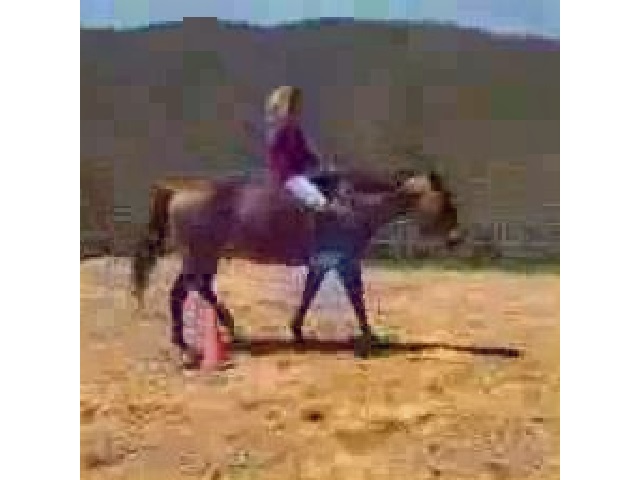}
\includegraphics[width=0.115\linewidth,height=0.115\linewidth,trim={3cm 0cm 3cm 0cm},clip]{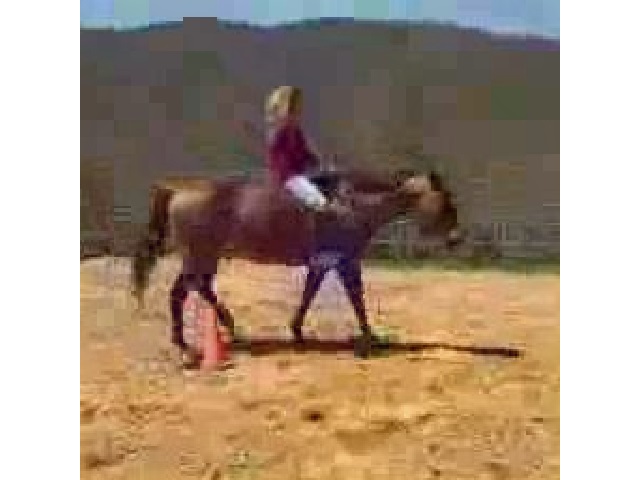}
\hspace*{0.5em}
\includegraphics[width=0.115\linewidth,height=0.115\linewidth,trim={0.5cm 0.4cm 1.4cm 1cm},clip]{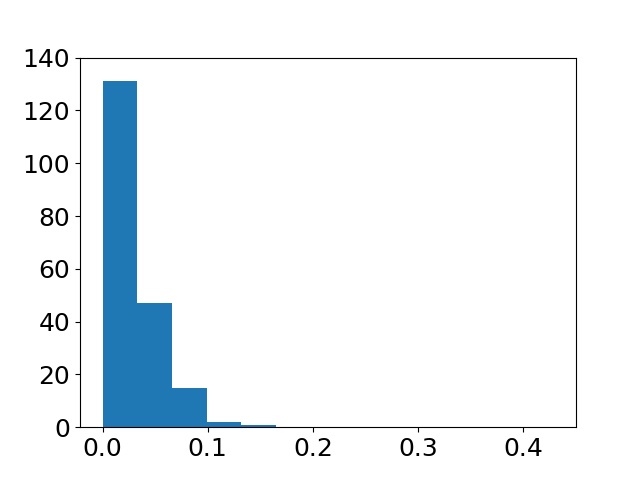} \\
 $k=0$ \hspace*{3.5em} $k=100$ \hspace*{2.5em} $k=200$ \hspace*{2.5em} $k=300$ \hspace*{2.5em} $k=400$ \hspace*{2.5em} $k=500$
\caption{Visualizations for a VGG16 network fine-tuned on UCF101. The middle columns display the saliency map over the same video frame of the action \textit{HorseRiding} while incrementally switching off the most $k$ relevant/salient neurons ($k=0, 100, 200, \dots, 500$) in the \textit{fc6} layer at test time. Excitation Dropout shows more robustness when more neurons are switched off. This is demonstrated through its ability to recover more of the saliency map even when a high percentage of the most salient neurons is dropped-out. This ability reflects the alternative learnt paths. Histograms of the leftmost and rightmost saliency maps are presented to demonstrate that Excitation Dropout has a wider range of saliency values.}
\label{fig:demo}
\end{figure*}

\subsection{Setup and Results: Generalization}
\label{sec:Performance}
We start by comparing Excitation Dropout with popular variants of dropout: Adaptive Dropout \cite{ba2013adaptive}, Information Dropout \cite{achille2018information}, and Curriculum Dropout \cite{morerio2017curriculum}. Adaptive Dropout drops neurons with low activations during training, Information Dropout is another data-dependent dropout approach, and Curriculum Dropout adjusts the dropout rate over time selecting neurons uniformly at random. We train a CNN-2 model from scratch on the datasets: Cifar10, Cifar100, Caltech256; We fine-tune an AlexNet pre-trained on ImageNet for the UCF101 dataset. We perform dropout in the first fully-connected layer of the networks (\textit{fc1} for CNN-2 and \textit{fc6} for AlexNet and VGGs) for Adaptive, Information, Standard, Curriculum, and Excitation Dropout. For Curriculum Dropout we fix the parameter $\gamma$ to $5*10^{-4}$ as in \cite{morerio2017curriculum}. Table \ref{table:analysis} reports classification accuracy (averaged over five trained models) over four benchmark datasets. Excitation Dropout has a higher accuracy compared to Adaptive, Information, and Curriculum Dropout. Curriculum Dropout is the second runner-up, and therefore we use it for comparison in the following extensive experimental analysis.

Fig.~\ref{fig:from_scratch} depicts the test accuracies over training iterations for the three datasets averaged over five trained models. After convergence, Excitation Dropout demonstrates a significant improvement in performance compared to other methods. We hypothesize that Excitation Dropout takes longer to converge due to the additional loop (Steps 2-4 in Fig.~\ref{fig:pipeline}) introduced in the learning process, and due to the learning of the alternative paths. We note that Excitation Dropout, during training, uses a different binary mask for each image in a minibatch, while in Standard Dropout, one random mask is employed per minibatch. To prove that it is precisely the fact that masks reflective of the particular input give rise to a boost in accuracy, and not the fact that different masks are used for different images, we add a comparison with Standard Dropout having a different random mask for each image. We refer to this accuracy as `Standard Dropout + Mask/Img' in the plots. As expected, the latter approach is comparable to Standard Dropout in performance.

Next, we evaluate the effectiveness of Excitation Dropout on popular network architectures that employ dropout layers: AlexNet, VGG16, VGG19. This is done by fine-tuning on the video recognition dataset UCF101. Fig.~\ref{fig:from_scratch} shows superior Excitation Dropout performance on AlexNet fine-tuned on UCF101. Table~\ref{table:fine_tuning} presents more comparative results on other deep architectures by reporting the accuracy after convergence. Again, Excitation Dropout demonstrates higher generalizability on the test data for all architectures. We also find that Excitation Dropout, applied in $fc6$ only, demonstrates higher performance compared to applying Standard Dropout in both $fc6$ and $fc7$ layers as in the original proposed architectures (AlexNet: 67.56\% \textit{vs.} 65.83\%, VGG16: 73.23\% \textit{vs.} 73.01\%, VGG19: 74.34\% \textit{vs.} 73.40\%).

In all experiments thus far, we set $p=0.5$ for Standard Dropout and $P=0.5$ for the base retaining probability of Excitation Dropout for fair comparison. For completeness, we add a sensitivity analysis of the parameters $p$ and $P$ in Table \ref{table:sensitivity}. Among the different dropout rates, the lowest accuracy for Excitation Dropout is greater than the highest accuracy of Standard Dropout for most datasets.

Excitation Dropout is a generic formulation that can be applied to any neural network layer. For a convolutional layer, a generic convolutional activation map is in the form of $[w,h,N]$, where $N$ is the number of feature maps while $w$ and $h$ are the spatial dimensions. To apply Excitation Dropout to a convolutional layer, first $p_{EB}$ is computed for each feature map $N$ as the sum of $p_{EB}$ across spatial locations $w$ and $h$. Specific 2D feature maps are then dropped-out following Eqn.~\ref{eq:EBdrop}. We test this on CNN-2 for Cifar-10. We observe an improvement respect to No Dropout ($76.91\%$), but consistent with \cite{Hintondrop2012,srivastava2014dropout}, the improvement is not as large as using dropout in fully connected layers (Excitation Dropout at $conv3$ $78.01\%$ \textit{vs.} Excitation Dropout at $fc1$ $81.94\%$).

\subsection{Setup and Results: Utilization of Network Neurons} 

In this section we examine how Excitation Dropout expands the network's utilization of neurons through the learnt alternative paths for a certain task. 

\cite{mittal2018recovering} introduced scoring functions to rank the filters in specific network layers including the \textit{average percentage of zero activations}, a metric to count how many neurons have zero activations, and the \textit{entropy of activations}, a metric to measure how much information is contained in the neurons of a layer. We analogously compute the \textit{Neurons ON}, which is the average number of non-zero activations, and the \textit{entropy of $p_{EB}$} which is higher when the probability distribution is spread out over more neurons in a layer. We also compute the \textit{peak $p_{EB}$}, which is expected to be lower on a more spread distribution. Moreover, \cite{ma2017less} introduced \textit{conservative filters}: filters whose parameters do not change significantly during training. Conservative filters reduce the effective number of parameters in a CNN and may limit the CNN’s modeling capacity for the target task. A conservative filter is a filter $k$ in layer $n$ whose weights have changed by $\Delta_{n}^k=\|\hat{w}_n^k-{w}_n^k\|$, where $\Delta_{n}^k$ is less than a threshold $\Delta$ (empirically set).

We evaluate these metrics for Excitation Dropout and compare against Standard and Curriculum Dropout in Table~\ref{table:metrics}. This is done on the same datasets and architectures considered in Section~\ref{sec:Performance}. All metrics are computed for the first fully-connected layer of the CNN-2 and VGG16 nets consisting of $2048$ and $4096$ neurons, respectively. We compute each metric over the test set of each dataset. Excitation Dropout consistently outperforms Standard and Curriculum Dropout in all the metrics over all datasets. Excitation Dropout shows a higher number of active neurons, a higher entropy over activations, and a probability distribution $p_{EB}$ that is more spread (higher entropy over $p_{EB}$) among the neurons of the layer, leading to a lower peak probability of $p_{EB}$ and therefore less specialized neurons. Averaging models having less specialized neurons results in higher robustness to information loss. We also observe a significantly smaller number of conservative filters when using Excitation Dropout. Fewer filters remain unchanged, \ie~do not sufficiently learn anything far from the random initialization. These results show that the models trained with Excitation Dropout were trained to be more informative, \ie~the contribution for the classification task is provided by a higher number of network neurons, reflecting the alternative learnt paths. 

Finally we report an extended analysis of the metrics: \textit{\# Neurons ON}, \textit{Peak $p_{EB}$}, \textit{Entropy of Activations}, and \textit{Entropy of $p_{EB}$} during training. Excitation Dropout shows a higher number of active neurons, a higher entropy over activations, and a probability distribution $p_{EB}$ that is more spread (higher entropy over $p_{EB}$) among the neurons of the layer, leading to a lower peak probability of $p_{EB}$ and therefore less specialized neurons. These results are observed to have consistent trends over all training iterations for all datasets. We present sample plots for Cifar100 (Fig.~\ref{fig:metrics_over_time_Cifar100}), a network trained from scratch, and UCF101 (Fig.~\ref{fig:metrics_over_time_UCF}), a fine-tuned network.



\subsection{Setup and Results: Resilience to Compression}

In this section, we simulate `Brain Damage' by dropping out neurons at test time. Fig.~\ref{fig:demo} demonstrates how a network utilizes the learnt alternative paths to capture the evidence of the class \textit{HorseRiding} in a video frame of the UCF101 dataset.
A VGG16 model is randomly selected from the five trained models used to report results in Table~\ref{table:fine_tuning}, and is fine-tuned once with each of Excitation, Curriculum, Standard, and No Dropout at the $fc6$ layer. We show the excitation saliency map obtained at the conv5-1 layer as we drop out a fixed number of the most relevant neurons from the layer in which dropout is applied at training time. A neuron is considered to be more relevant if it has a higher $p_{EB}$. In the first column of frames of Fig.~\ref{fig:demo}, the original saliency maps for the different models are shown. As already highlighted in Table~\ref{table:metrics}, the original saliency map obtained from the model trained with Excitation Dropout is more spread as compared to that of the other schemes, which present more pronounced red peaks. In the remaining columns of Fig.~\ref{fig:demo}, we present the saliency maps the model is able to maintain when the $100, 200, 300, 400, 500$ most relevant neurons are dropped-out. Despite the increasing number of relevant neurons being dropped-out, Excitation Dropout is capable of restoring more of the saliency map contributing to \textit{HorseRiding}. This means that the network with Excitation Dropout was trained to find alternative paths that belong to the same \textit{HorseRiding}-relevant cues of the image. Despite the fact that we are considering the \textit{worst-case} scenario, where we are switching off the \textit{most} relevant neurons at test time, Excitation Dropout shows the most robustness. The histograms of the leftmost and rightmost saliency maps show that Excitation Dropout has a wider range of saliency values. 

\begin{figure*}[h!]
    \centering
    Cifar10\\[0.2em]
    \includegraphics[width=0.325\textwidth,trim={0.2cm 0cm 0.7cm 0.5cm},clip]{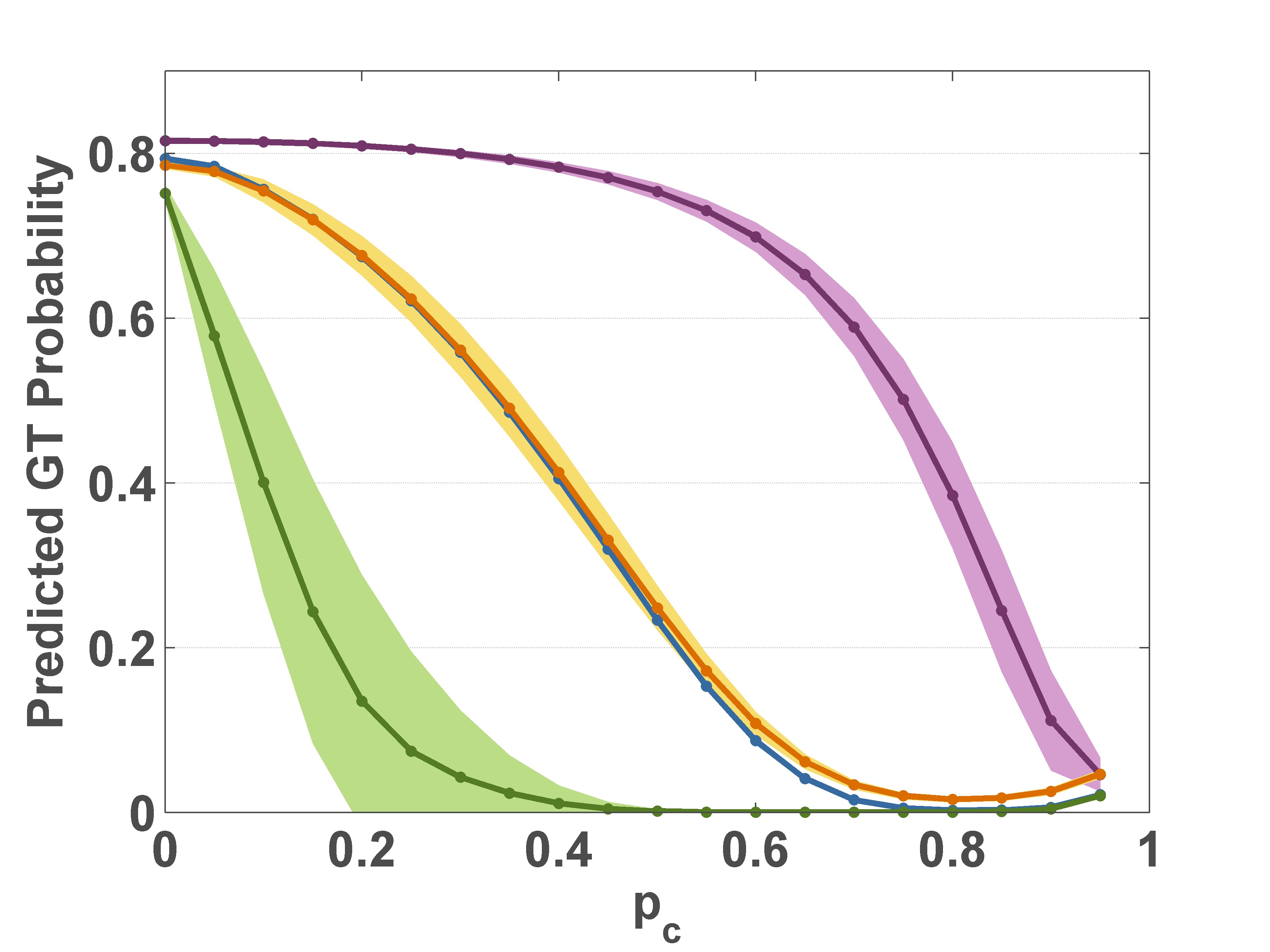}
    \includegraphics[width=0.325\textwidth,trim={0.2cm 0cm 0.7cm 0.5cm},clip]{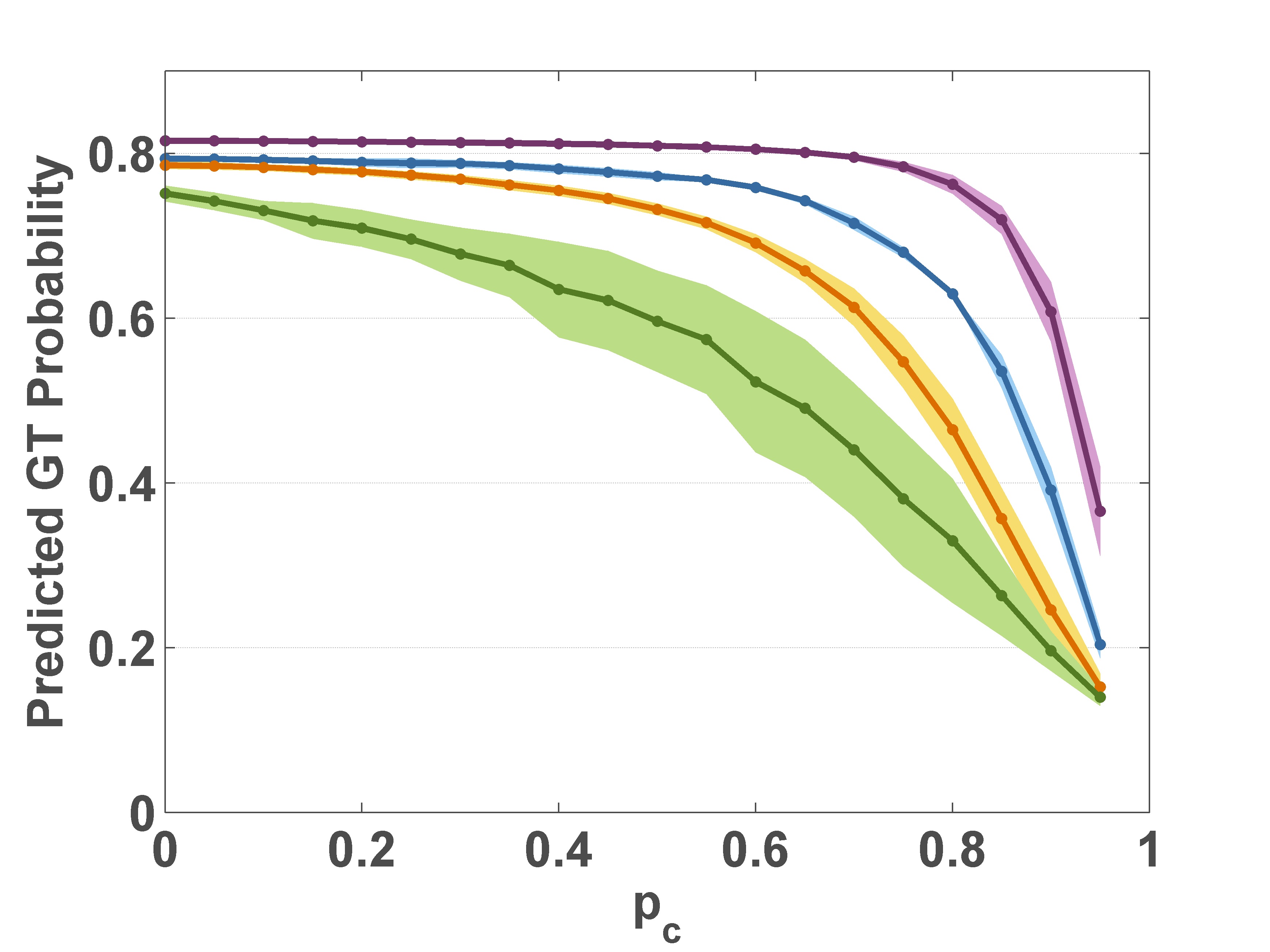}
        \includegraphics[width=0.325\textwidth,trim={0.2cm 0cm 0.6cm 0.5cm},clip]{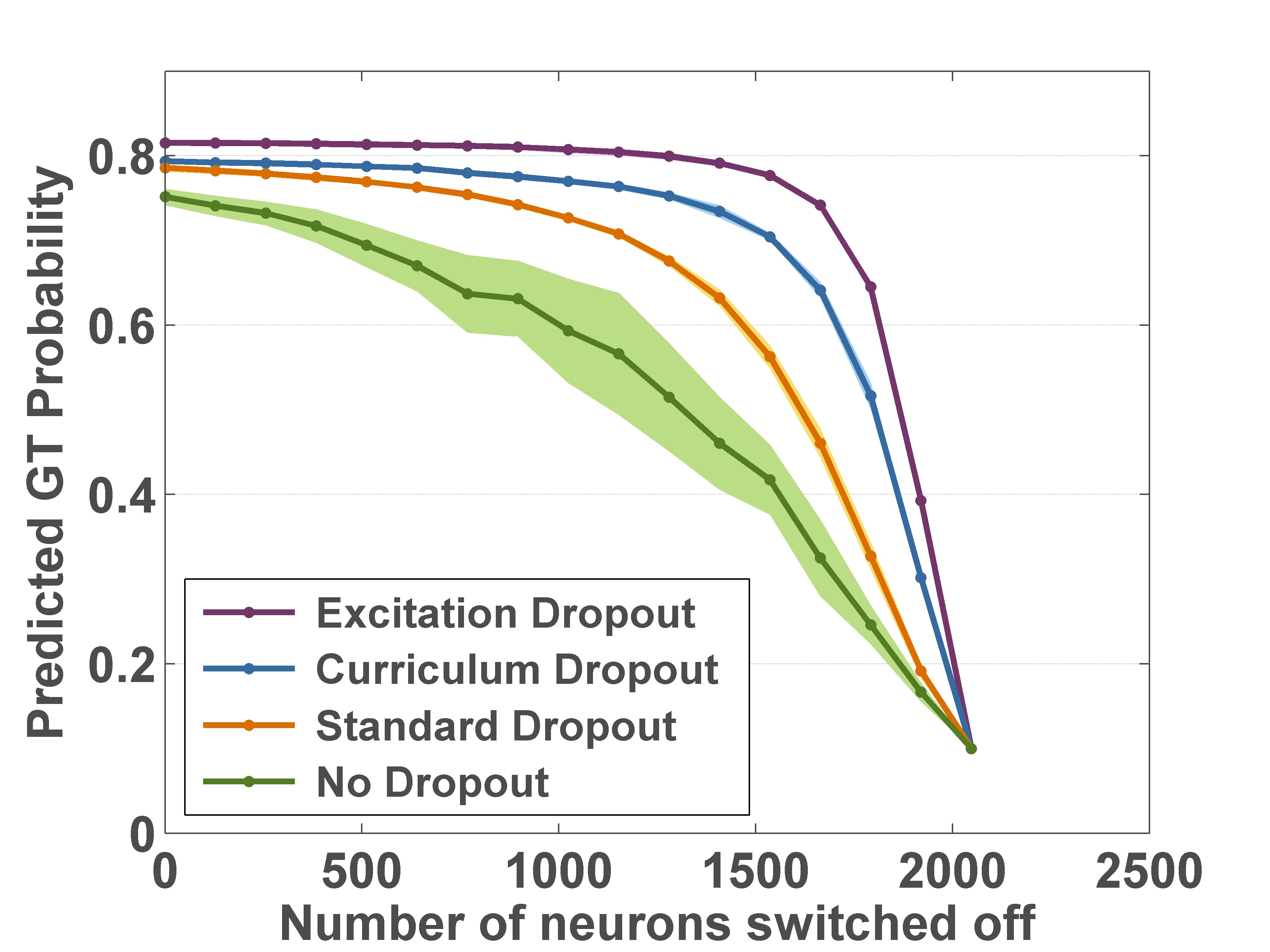}
    Cifar100\\[0.2em]
        \includegraphics[width=0.325\textwidth,trim={0.2cm 0cm 0.7cm 0.5cm},clip]{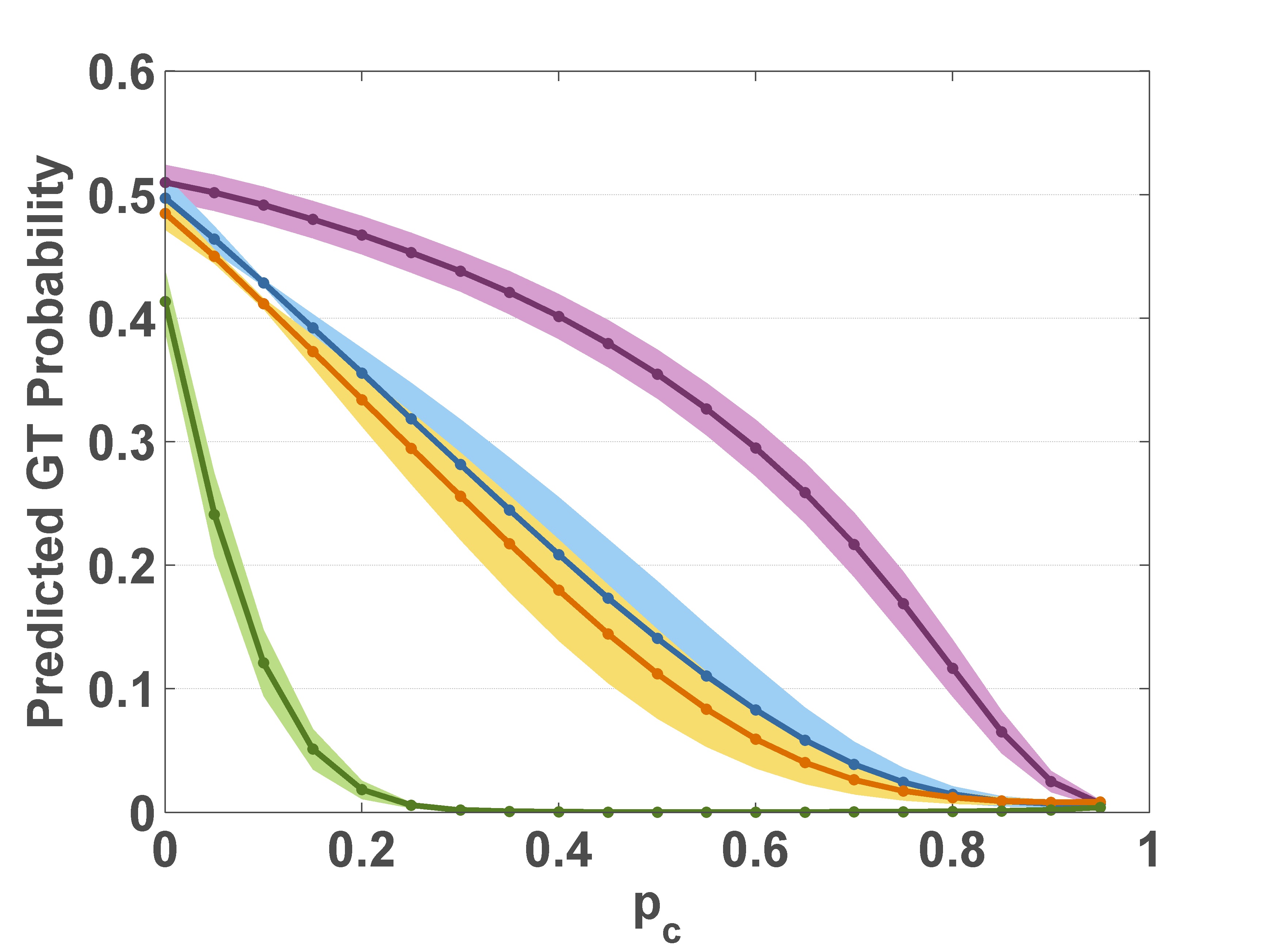}
        \includegraphics[width=0.325\textwidth,trim={0.2cm 0cm 0.7cm 0.5cm},clip]{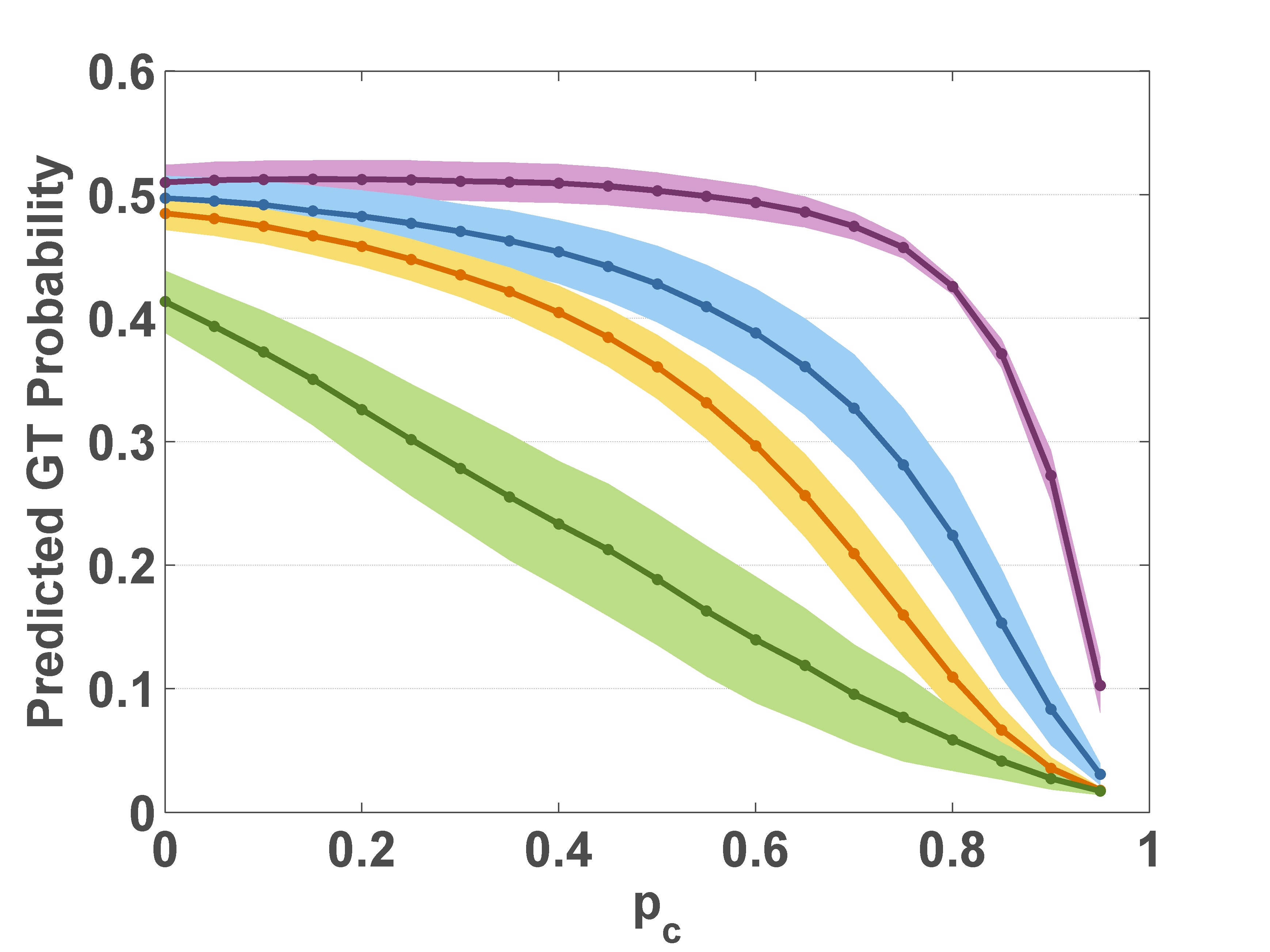}
        \includegraphics[width=0.325\textwidth,trim={0.2cm 0cm 0.6cm 0.5cm},clip]{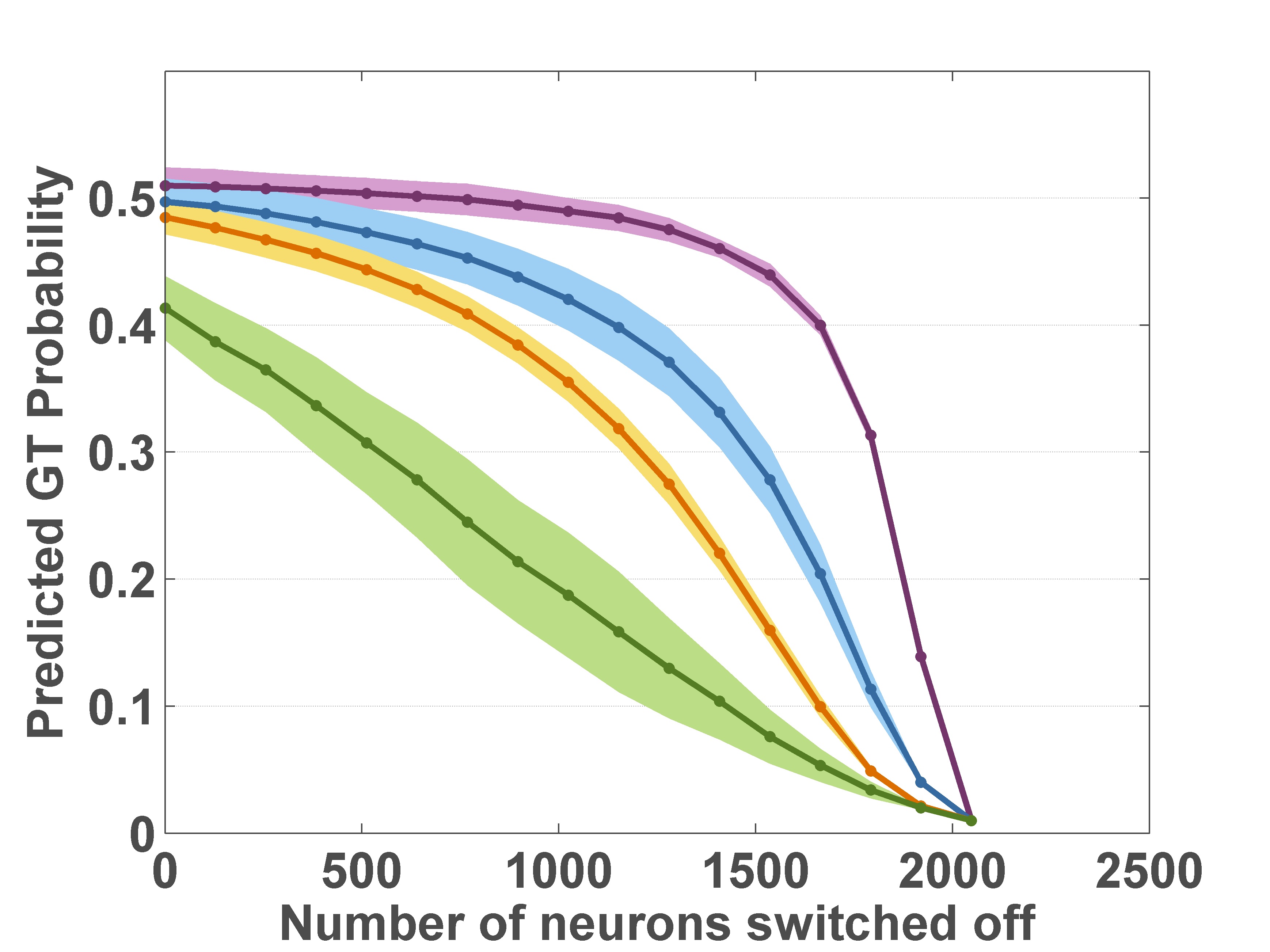}
        Caltech256\\[0.2em]      
        \includegraphics[width=0.325\textwidth,trim={0.2cm 0cm 0.7cm 0.5cm},clip]{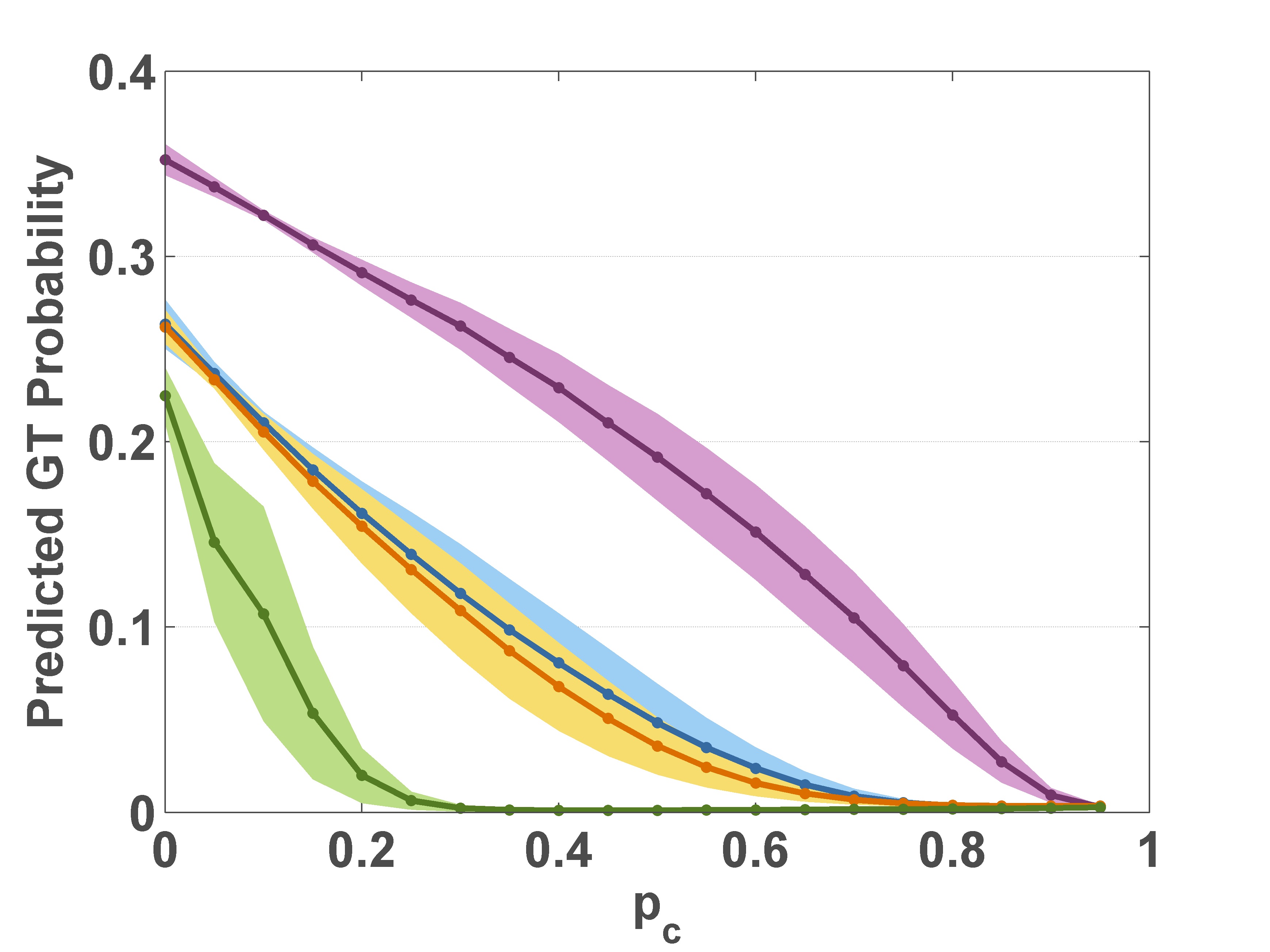}
        \includegraphics[width=0.325\textwidth,trim={0.2cm 0cm 0.7cm 0.5cm},clip]{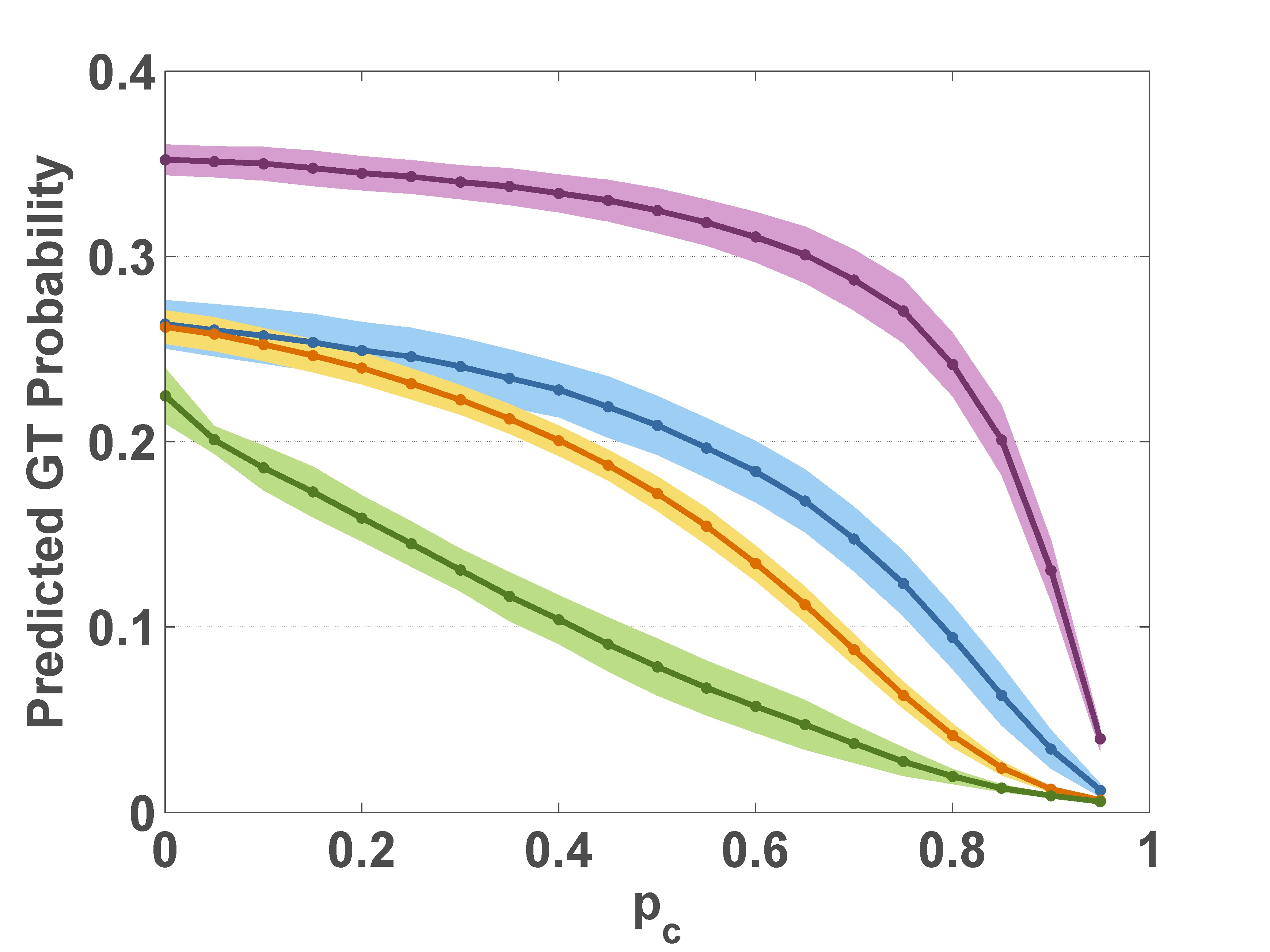}
        \includegraphics[width=0.325\textwidth,trim={0.2cm 0cm 0.6cm 0.5cm},clip]{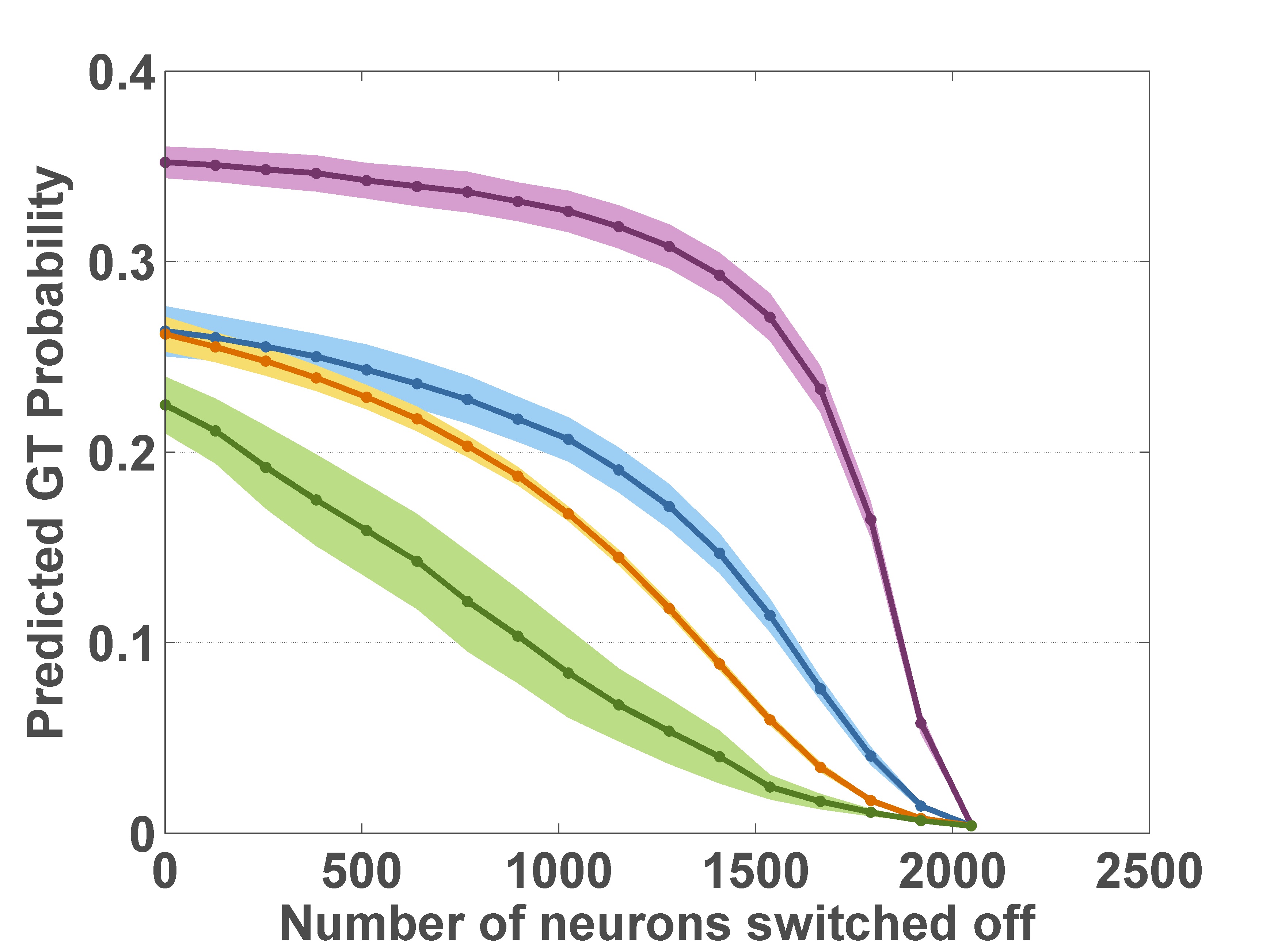}
        UCF101\\[0.2em]
    \includegraphics[width=0.325\textwidth,height=0.26\textwidth,trim={0.2cm 0cm 0.7cm 0.5cm},clip]{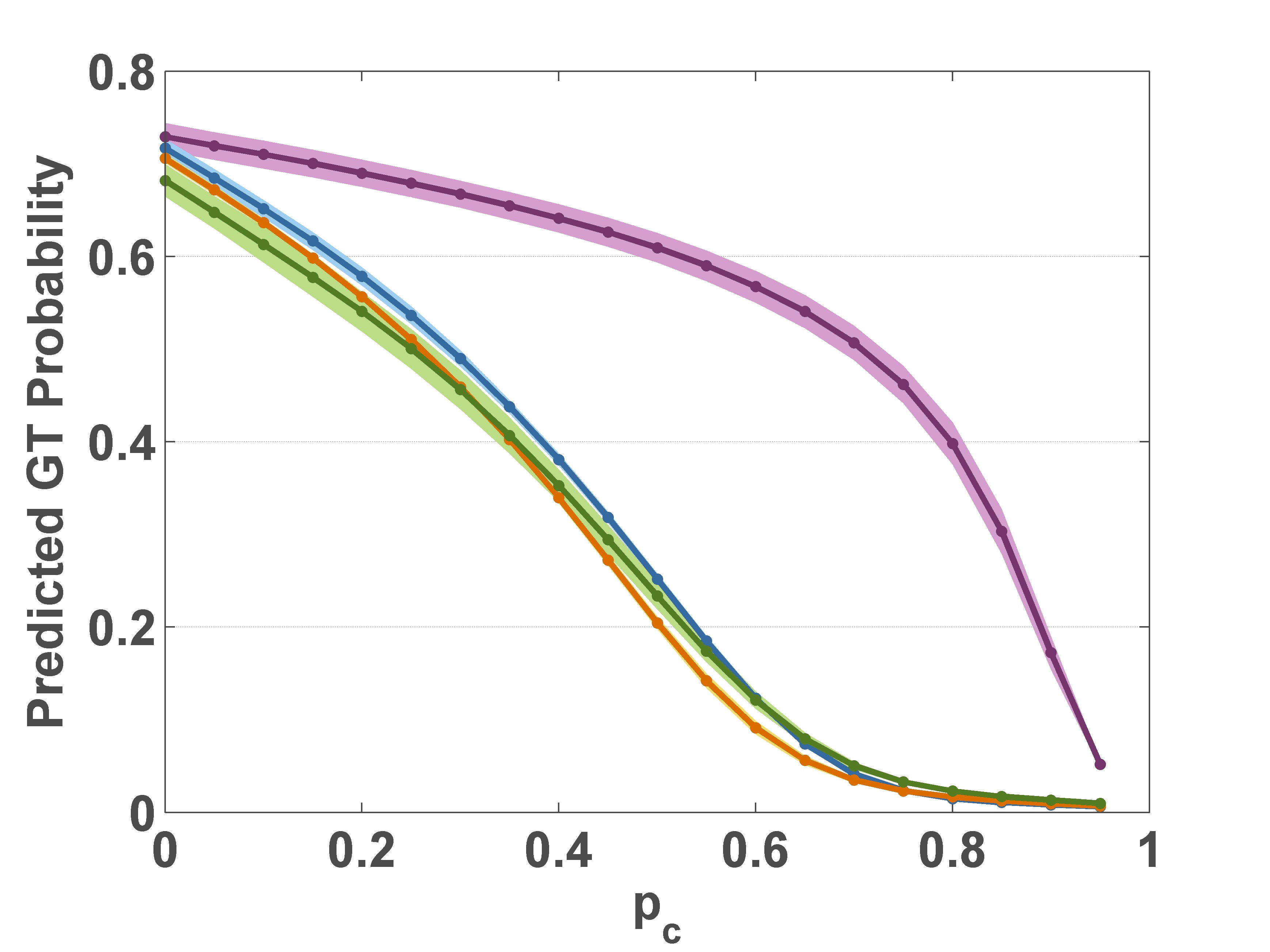}
        \includegraphics[width=0.325\textwidth,height=0.26\textwidth,trim={0.2cm 0cm 0.7cm 0.5cm},clip]{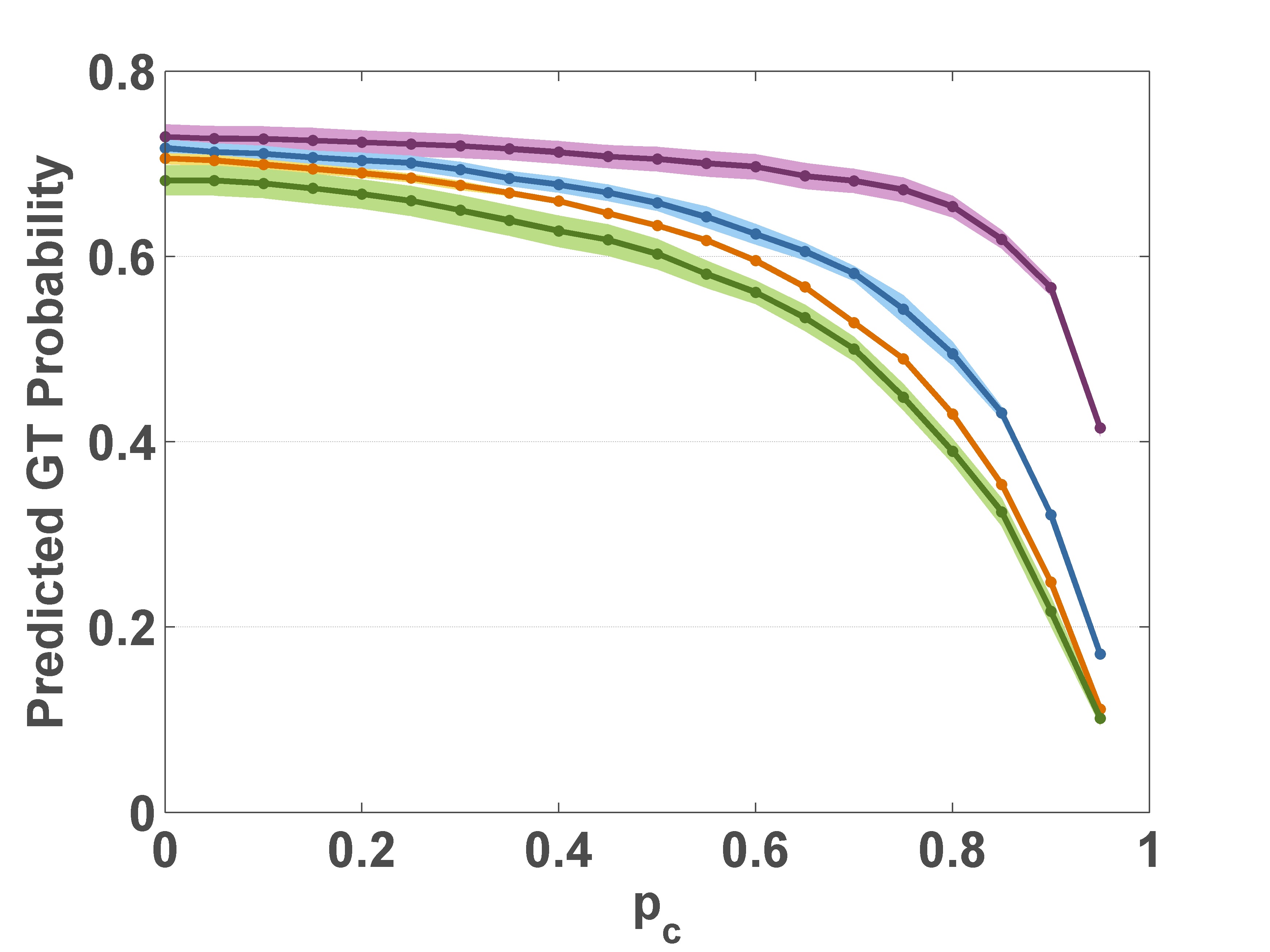}
        \includegraphics[width=0.325\textwidth,height=0.26\textwidth,trim={0.2cm 0cm 0.7cm 0.5cm},clip]{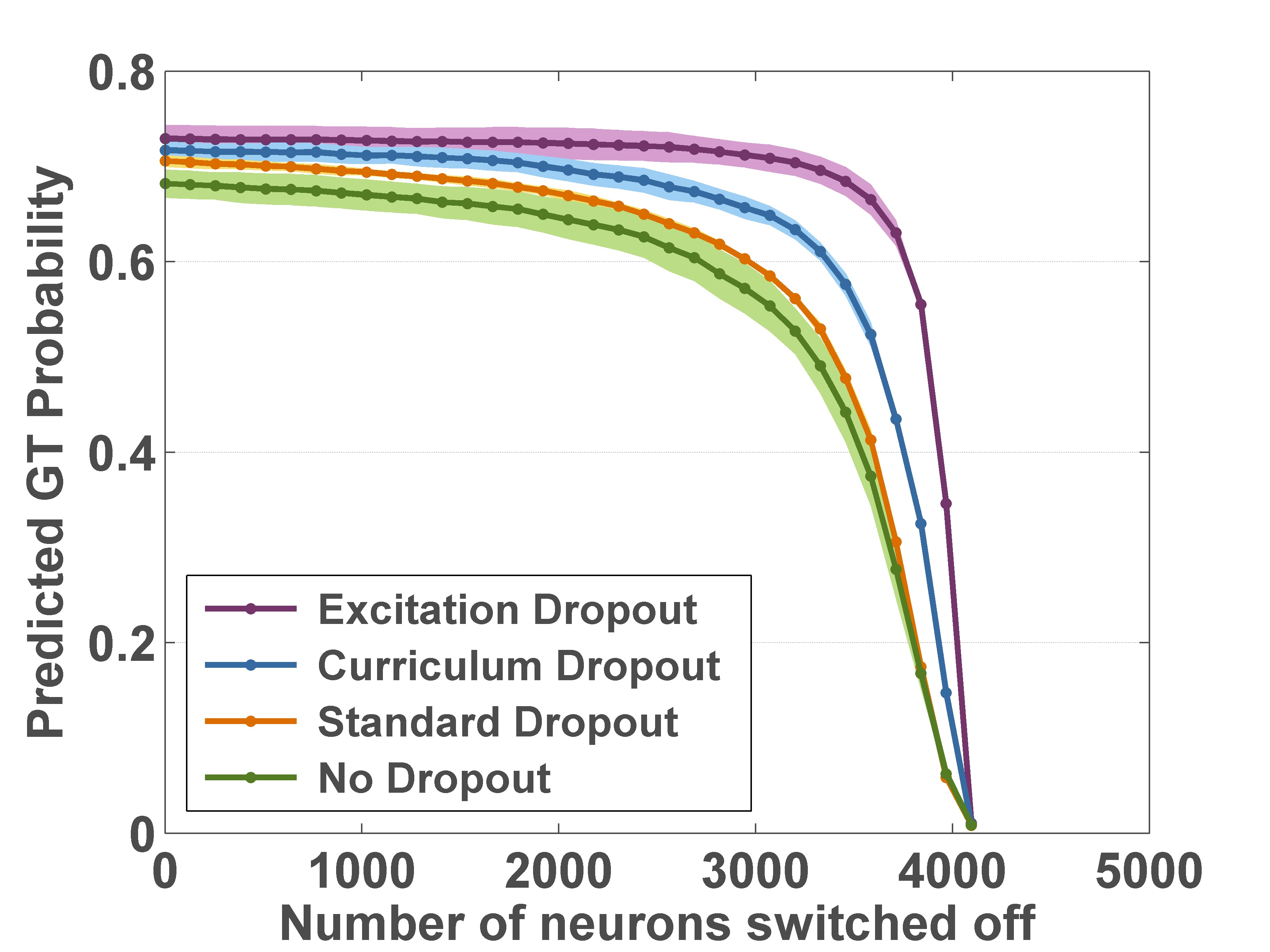}\vspace{1 em}
\caption{Robustness of predicted ground-truth class probabilities as more neurons are dropped out for each dataset's test images. For Cifar10, Cifar100 and Caltech256 we train CNN-2 from scratch with Excitation, Curriculum, Standard, and No Dropout at the $fc1$ layer. For UCF101 we fine-tune VGG16 with Excitation, Curriculum, Standard, and No Dropout at the $fc6$ layer. For both architectures we average results over five trained models. The standard deviation is depicted around the mean curve using a lighter shade. Left: the most relevant neurons with respect to the $p_c$ threshold are switched off. Center: the least relevant neurons with respect to the $p_c$ threshold are switched off. Right: $k$ neurons are randomly switched off. In all scenarios, Excitation Dropout shows more robustness to network compression (dropping $fc$ neurons $\equiv$ removing filters).}
\label{fig:comp_UCF}
\end{figure*}

While Fig.~\ref{fig:demo} visualizes one example qualitatively, Fig.~\ref{fig:comp_UCF} presents a complete quantitative analysis on the entire test set after training is complete. We study how the predicted ground-truth (GT) probability changes as more neurons are dropped-out at test time. On the left we present the worst case when the neurons dropped are the most relevant to the prediction. The horizontal axis in the graph represents $p_c$, where $0 \le p_c \le 1$ is the cumulative sum of $p_{EB}$ of neurons which will be switched off starting from the most `important.' The analysis is performed for $p_c=\{0, 0.05, \dots, 0.90, 0.95\}$. In the center, we present an analogous analysis starting to drop from the `least' relevant neurons. On the right, we present the random case (more realistic) when $k$ neurons ($k=0, 128, 256, \dots, 4096$) for VGG16, when $k$ neurons ($k=0, 128, 256, \dots, 2048$) for CNN-2 are randomly switched off. As we drop more neurons, Excitation Dropout (purple curves) is capable of maintaining a much less steep decline of GT probability, indicating more robustness against network compression. We obtain a similar behaviour for all the considered datasets.

\begin{table*}[t]
\centering
\begin{tabular}{ll|c|c|c|c} 
\hline
& & \makecell{\textbf{Standard} \\ \textbf{Dropout \scriptsize{\textit{first fc}}}}
& \makecell{\textbf{Standard} \\ \textbf{Dropout \scriptsize{\textit{both fcs}}}}
& \makecell{\textbf{Activation} \\ \textbf{Dropout}} & \makecell{\textbf{Excitation} \\ \textbf{Dropout} } \\ 
\hline
\multirow{4}{*}{\textbf{Dataset}} & \textbf{\textit{Cifar10}} & 80.13\% & 80.80\%
& 81.33\%  & \bf{81.94}\%  \\
& \textbf{\textit{Cifar100}} & 50.36\% & 51.96\%
& 51.43\%  & \bf{52.04}\%  \\
& \textbf{\textit{Caltech256}} & 28.73\% & 31.77\%
& 35.29\%  & \bf{35.77}\%  \\
& \textbf{\textit{UCF101}}  & 71.93\%  & 73.01\%
& 72.56\%  & \bf{73.23}\% \\
\hline
\multirow{2}{*}{\textbf{Runtime}}&\textbf{CNN-2 \textit{(1 iter)}}  & \bf{0.152 $\pm$ 0.007 } & 0.161 $\pm$ 0.006
& 0.184 $\pm$ 0.006  & 0.189 $\pm$ 0.007 \\
&{\textbf{VGG16 \textit{(1 iter)}}}  & \bf{2.279 $\pm$ 0.028} & 2.290 $\pm$ 0.018
& 2.803 $\pm$ 0.024  & 2.820 $\pm$ 0.031 \\
\hline
\end{tabular}
\vspace*{0.2em}
\caption{Accuracy and runtime comparison between Standard, Activation and Excitation Dropout: Average training time of 100 iterations (in seconds, batch size=50) for a Caffe python layer on a GTX Titan X GPU and Intel(R) Xeon(R) CPU E5-2650 v3 @ 2.30GHz.}
\label{table:runtime_analysis}
\vspace*{-1em}
\end{table*}

\section{Activation Dropout: A lighter version}
\label{act}

Popular dropout methods (\eg~Adaptive Dropout) drop \textit{useless} neurons with low activations. So far we have demonstrated that dropping neurons based on their top-down attention brings benefits. We now introduce a lighter version of Excitation Dropout, \textit{Activation Dropout}, where, in contrast with Excitation Dropout, the forward activations only are used to determine dropout probabilities. As in the Excitation Dropout formulation, the forward activation of neuron ${a}_j$ is defined as $\widehat{a}_j=\phi(\sum_iw_{ij}\widehat{a}_i+b_i)$, where $N$ is the number of neurons in the specific layer in which dropout will be applied. For Activation Dropout, we normalize the activations as follows: $p_{Act}(a_j)=\widehat{a}_j/\sum_{i\in {1 \dots N}}\widehat{a}_i$. We now define the retaining probability $p$ of Activation Dropout as:
\begin{equation}
\label{eq:ACTdrop}
p=1-\frac{(1-P)*(N-1)*p_{Act}}{((1-P)*N-1)*p_{Act}+P}
\end{equation}
where $p_{Act}$ is the probability distribution computed over the activations of the layer in which we apply dropout. While Excitation Dropout utilizes top-down attention, a function of the forward and backward passes, Activation Dropout is a computationally simpler measure that only utilizes the forward activations. The top part of Table \ref{table:runtime_analysis} presents a comparison of classification accuracy between the two proposed dropout variants applied at the first \textit{fc} layer. The bottom part of Table \ref{table:runtime_analysis} presents a run-time analysis for Excitation, Activation and Standard Dropout, for the two main architectures used in this work. Excitation Dropout consistently outperforms Standard Dropout, with zero increase in test-time computational complexity. In training, there is a moderate increase in computation: in the worst case, Excitation Dropout will take double (same O-notation complexity) the training time of Standard Dropout. This will happen when the utilized Excitation Dropout maps are at the first layer of the network. If a middle layer is used, Excitation Dropout requires only a \textit{partial} additional forward-backward pass. We apply dropout to \textit{fc} layers close to the end of the network to reduce this overhead.

An efficiency-accuracy trade-off exists between Activation Dropout and Excitation Dropout. The latter utilizes the backward pass as additional information for determining the contribution of a neuron to a model's prediction, hence the added accuracy. The former instead, only utilizes the forward pass, hence the higher efficiency. Therefore, $p_{Act}$ can be used as a rough approximation of $p_{EB}$ when efficiency is a priority over a slight compromise in accuracy. 

We now apply the lighter Activation Dropout to (a) the more modern WideResNet architecture, and (b) the larger-scale ImageNet dataset. First, we replace the dropout of WideResNet (WRN-28-10) by our lighter version Activation Dropout obtaining on Cifar10: 3.88\% test error (compared to 4.17\% \cite{zagoruyko2016wide}). We use the default depth and width architecture parameters of 28 and 10 respectively, starting learning rate of $10^{-1}$ with a decay ratio of 0.2, and batch size of 128. Second, we train an AlexNet from scratch on ImageNet, once using Standard Dropout and another using our Activation Dropout, both applied to $fc6$ layer. Using basic training with an Adam optimizer, batch size of 50, and a learning rate of $10^{-5}$, Activation Dropout obtained a test accuracy of 48.54\% while Standard Dropout achieved a test accuracy of 44.43\%. This further demonstrates the generalization ability of our proposed dropout technique.



\section{Conclusion}
\label{conclus}
We propose a new regularization scheme that encourages the learning of alternative paths in a neural network by deliberately paralyzing high-saliency neurons that contribute more to a network's prediction during training. In experiments on four image/video recognition datasets, and on different architectures, we demonstrate that our approach yields better generalization on unseen data, higher utilization of network neurons, and higher resilience to network compression.

\section*{Acknowledgements}
This work was supported in part by the Defense Advanced Research Projects Agency (DARPA) Explainable Artificial Intelligence (XAI) program, an IBM PhD Fellowship, a Hariri Graduate Fellowship, and gifts from Adobe and NVidia. The U.S. Government is authorized to reproduce and distribute reprints for Governmental purposes notwithstanding any copyright annotation thereon. Disclaimer: The views and conclusions contained herein are those of the authors and should not be interpreted as necessarily representing the official policies or endorsements, either expressed or implied, of DARPA, DOI/IBC, or the U.S. Government.

\bibliographystyle{spbasic}      
\bibliography{fonti}

\begin{thebibliography}{34}
\providecommand{\natexlab}[1]{#1}
\providecommand{\url}[1]{{#1}}
\providecommand{\urlprefix}{URL }
\expandafter\ifx\csname urlstyle\endcsname\relax
  \providecommand{\doi}[1]{DOI~\discretionary{}{}{}#1}\else
  \providecommand{\doi}{DOI~\discretionary{}{}{}\begingroup
  \urlstyle{rm}\Url}\fi
\providecommand{\eprint}[2][]{\url{#2}}

\bibitem[{Achille and Soatto(2018)}]{achille2018information}
Achille A, Soatto S (2018) Information dropout: Learning optimal
  representations through noisy computation. IEEE Transactions on Pattern
  Analysis and Machine Intelligence (PAMI)

\bibitem[{Ba and Frey(2013)}]{ba2013adaptive}
Ba J, Frey B (2013) Adaptive dropout for training deep neural networks. In:
  Advances in Neural Information Processing Systems (NIPS)

\bibitem[{Baldi and Sadowski(2013)}]{Baldi2013}
Baldi P, Sadowski PJ (2013) Understanding dropout. Advances in Neural
  Information Processing Systems (NIPS)

\bibitem[{Deng et~al.(2009)Deng, Dong, Socher, Li, Li, and
  Fei-Fei}]{deng2009imagenet}
Deng J, Dong W, Socher R, Li LJ, Li K, Fei-Fei L (2009) Imagenet: A large-scale
  hierarchical image database. In: Proc.\ IEEE Conference on Computer Vision
  and Pattern Recognition (CVPR)

\bibitem[{Gal et~al.(2017)Gal, Hron, and Kendall}]{gal2017concrete}
Gal Y, Hron J, Kendall A (2017) Concrete dropout. In: Advances in Neural
  Information Processing Systems (NIPS)

\bibitem[{Ghiasi et~al.(2018)Ghiasi, Lin, and Le}]{ghiasi2018dropblock}
Ghiasi G, Lin TY, Le QV (2018) Dropblock: A regularization method for
  convolutional networks. In: Advances in Neural Information Processing Systems
  (NIPS)

\bibitem[{Gomez et~al.(2018)Gomez, Zhang, Swersky, Gal, and
  Hinton}]{gomez2018targeted}
Gomez AN, Zhang I, Swersky K, Gal Y, Hinton GE (2018) Targeted dropout. In:
  NIPS Compact Deep Neural Network Representation with Industrial Applications
  Workshop

\bibitem[{Griffin et~al.(2007)Griffin, Holub, and Perona}]{griffinHolubPerona}
Griffin G, Holub A, Perona P (2007) Caltech-256 object category dataset. Tech.
  Rep. 7694, California Institute of Technology,
  \urlprefix\url{http://authors.library.caltech.edu/7694}

\bibitem[{Hebb(2005)}]{hebb2005organization}
Hebb DO (2005) The organization of behavior: A neuropsychological theory.
  Psychology Press

\bibitem[{Hinton et~al.(2015)Hinton, Vinyals, and Dean}]{hinton2015distill}
Hinton G, Vinyals O, Dean J (2015) Distilling the knowledge in a neural
  network. In: NIPS Deep Learning and Representation Learning Workshop

\bibitem[{Hinton et~al.(2012)Hinton, Srivastava, Krizhevsky, Sutskever, and
  Salakhutdinov}]{Hintondrop2012}
Hinton GE, Srivastava N, Krizhevsky A, Sutskever I, Salakhutdinov R (2012)
  Improving neural networks by preventing co-adaptation of feature detectors.
  CoRR abs/1207.0580

\bibitem[{Kang et~al.(2017)Kang, Li, and Tao}]{kang2017shakeout}
Kang G, Li J, Tao D (2017) Shakeout: A new approach to regularized deep neural
  network training. IEEE Transactions on Pattern Analysis and Machine
  Intelligence (PAMI)

\bibitem[{Kingma et~al.(2015)Kingma, Salimans, and
  Welling}]{kingma2015variational}
Kingma DP, Salimans T, Welling M (2015) Variational dropout and the local
  reparameterization trick. In: Advances in Neural Information Processing
  Systems (NIPS)

\bibitem[{Krizhevsky et~al.(2009)Krizhevsky, Hinton
  et~al.}]{krizhevsky2009learning}
Krizhevsky A, Hinton G, et~al. (2009) Learning multiple layers of features from
  tiny images. Citeseer

\bibitem[{Krizhevsky et~al.(2012)Krizhevsky, Sutskever, and
  Hinton}]{krizhevsky2012imagenet}
Krizhevsky A, Sutskever I, Hinton GE (2012) Imagenet classification with deep
  convolutional neural networks. In: Advances in Neural Information Processing
  Systems (NIPS)

\bibitem[{Li et~al.(2016)Li, Gong, and Yang}]{li2016improved}
Li Z, Gong B, Yang T (2016) Improved dropout for shallow and deep learning. In:
  Advances in Neural Information Processing Systems (NIPS)

\bibitem[{Ma et~al.(2017)Ma, Bargal, Zhang, Sigal, and Sclaroff}]{ma2017less}
Ma S, Bargal SA, Zhang J, Sigal L, Sclaroff S (2017) Do less and achieve more:
  Training {CNN}s for action recognition utilizing action images from the web.
  {Pattern Recognition}

\bibitem[{Miconi et~al.(2018)Miconi, Clune, and
  Stanley}]{miconi2018differentiable}
Miconi T, Clune J, Stanley KO (2018) Differentiable plasticity: training
  plastic neural networks with backpropagation. arXiv preprint arXiv:180402464

\bibitem[{Mittal et~al.(2018)Mittal, Bhardwaj, Khapra, and
  Ravindran}]{mittal2018recovering}
Mittal D, Bhardwaj S, Khapra MM, Ravindran B (2018) Recovering from random
  pruning: On the plasticity of deep convolutional neural networks. Winter
  Conference on Applications of Computer Vision

\bibitem[{Morerio et~al.(2017)Morerio, Cavazza, Volpi, Vidal, and
  Murino}]{morerio2017curriculum}
Morerio P, Cavazza J, Volpi R, Vidal R, Murino V (2017) Curriculum dropout. In:
  Proc.\ IEEE International Conference on Computer Vision (ICCV)

\bibitem[{Rennie et~al.(2014)Rennie, Goel, and Thomas}]{rennie2014annealed}
Rennie SJ, Goel V, Thomas S (2014) Annealed dropout training of deep networks.
  In: Spoken Language Technology Workshop (SLT), 2014 IEEE, IEEE, pp 159--164

\bibitem[{Selvaraju et~al.(2017)Selvaraju, Cogswell, Das, Vedantam, Parikh, and
  Batra}]{gradcam}
Selvaraju RR, Cogswell M, Das A, Vedantam R, Parikh D, Batra D (2017) Grad-cam:
  Visual explanations from deep networks via gradient-based localization. In:
  Proc.\ IEEE International Conference on Computer Vision (ICCV)

\bibitem[{Simonyan and Zisserman(2014)}]{simonyan2014very}
Simonyan K, Zisserman A (2014) Very deep convolutional networks for large-scale
  image recognition. arXiv preprint arXiv:14091556

\bibitem[{Song et~al.(2000)Song, Miller, and Abbott}]{song2000competitive}
Song S, Miller KD, Abbott LF (2000) Competitive hebbian learning through
  spike-timing-dependent synaptic plasticity. Nature neuroscience 3(9):919

\bibitem[{Soomro et~al.(2012)Soomro, Zamir, and Shah}]{soomro2012ucf101}
Soomro K, Zamir AR, Shah M (2012) {UCF101}: A dataset of 101 human actions
  classes from videos in the wild. arXiv preprint arXiv:12120402

\bibitem[{Srivastava et~al.(2014)Srivastava, Hinton, Krizhevsky, Sutskever, and
  Salakhutdinov}]{srivastava2014dropout}
Srivastava N, Hinton G, Krizhevsky A, Sutskever I, Salakhutdinov R (2014)
  Dropout: A simple way to prevent neural networks from overfitting. Journal of
  Machine Learning Research (JMLR) 15(1):1929--1958

\bibitem[{Wager et~al.(2013)Wager, Wang, and Liang}]{wager2013dropout}
Wager S, Wang S, Liang PS (2013) Dropout training as adaptive regularization.
  In: Advances in Neural Information Processing Systems (NIPS)

\bibitem[{Wan et~al.(2013)Wan, Zeiler, Zhang, Le~Cun, and
  Fergus}]{wan2013regularization}
Wan L, Zeiler M, Zhang S, Le~Cun Y, Fergus R (2013) Regularization of neural
  networks using dropconnect. In: Proc.\ International Conference on Machine
  Learning (ICML)

\bibitem[{Wang and Manning(2013)}]{wang2013fast}
Wang S, Manning C (2013) Fast dropout training. In: Proc.\ International
  Conference on Machine Learning (ICML), pp 118--126

\bibitem[{Wu and Gu(2015)}]{wu2015towards}
Wu H, Gu X (2015) Towards dropout training for convolutional neural networks.
  Neural Networks 71:1--10

\bibitem[{Zagoruyko and Komodakis(2016)}]{zagoruyko2016wide}
Zagoruyko S, Komodakis N (2016) Wide residual networks. Proc\ British Machine
  Vision Conference (BMVC)

\bibitem[{Zhang et~al.(2016)Zhang, Lin, Brandt, Shen, and
  Sclaroff}]{zhang2016top}
Zhang J, Lin Z, Brandt J, Shen X, Sclaroff S (2016) Top-down neural attention
  by excitation backprop. In: Proc.\ European Conference on Computer Vision
  (ECCV)

\bibitem[{Zhang et~al.(2017)Zhang, Bargal, Lin, Brandt, Shen, and
  Sclaroff}]{zhang2017top}
Zhang J, Bargal SA, Lin Z, Brandt J, Shen X, Sclaroff S (2017) Top-down neural
  attention by excitation backprop. International Journal of Computer Vision
  (IJCV) pp 1--19

\bibitem[{Zhou et~al.(2016)Zhou, Khosla, Lapedriza, Oliva, and Torralba}]{cam}
Zhou B, Khosla A, Lapedriza A, Oliva A, Torralba A (2016) Learning deep
  features for discriminative localization. In: Proc.\ IEEE Conference on
  Computer Vision and Pattern Recognition (CVPR)

\end{thebibliography}

\end{document}